\documentclass[Afour,sageh,times]{sagecustom}

\usepackage{moreverb,url}

\usepackage{mathtools}
\usepackage{amsmath}
\usepackage{amssymb}
\usepackage{bm}
\usepackage{balance}
\usepackage{siunitx}
\usepackage{acro}
\usepackage{cleveref}
\usepackage{mathrsfs}
\usepackage{tabularx}
\usepackage{threeparttable}
\usepackage{tikz}
\usepackage{tikzscale}
\usepackage[page]{appendix}

\DeclareAcronym{DOF}{short = DoF, long = degrees of freedom}
\DeclareAcronym{MAV}{short = MAV, long = micro aerial vehicle}
\DeclareAcronym{OMAV}{short = OMAV, long = omnidirectional micro aerial vehicle}
\DeclareAcronym{CAD}{short = CAD, long = computer-aided design}
\DeclareAcronym{IMU}{short = IMU, long = inertial measurement unit}
\DeclareAcronym{EKF}{short = EKF, long = extend\textbf{}ed Kalman filter}
\DeclareAcronym{LQRI}{short = LQRI, long = linear-quadratic regulator with integral action}
\DeclareAcronym{MPC}{short = MPC, long = model predictive controller}
\DeclareAcronym{PID}{short = PID, long = proportional-integral-derivative}
\DeclareAcronym{IQR}{short = IQR, long = interquartile range}
\DeclareAcronym{SG}{short = SG, long = Savitzky-Golay}
\DeclareAcronym{COM}{short = CoM, long = center of mass}

\renewcommand{\ac}[1]{#1}
\newcommand\BibTeX{{\rmfamily B\kern-.05em \textsc{i\kern-.025em b}\kern-.08em
T\kern-.1667em\lower.7ex\hbox{E}\kern-.125emX}}
\newcommand{\f}[1]{\ _{#1}}
\newcommand*\diff{\mathop{}\!\mathrm{d}}
\newcommand{\sign}{\mathrm{sign}}
\newcommand{\tr}{\mathrm{tr}}
\renewcommand{\vec}[1]{\bm{#1}}

\begin{document}

\runninghead{Allenspach, Bodie, Brunner, et al}

\title{Design and optimal control of a tiltrotor micro aerial vehicle for efficient omnidirectional flight}

\author{Mike Allenspach\affilnum{1}\affilnum{*}, Karen Bodie\affilnum{1}\affilnum{*}, Maximilian Brunner\affilnum{1}\affilnum{*}, Luca Rinsoz\affilnum{1}, Zachary Taylor\affilnum{1}, Mina Kamel\affilnum{1}, Roland Siegwart\affilnum{1} and Juan Nieto\affilnum{1}}

\affiliation{\affilnum{*}Authors contributed equally to the work \\ \affilnum{1}ETH Z\"{u}rich, Switzerland}

\corrauth{Karen Bodie, Maximilian Brunner\\
Autonomous Systems Lab, ETH Z\"{u}rich,\\
Leonhardstrasse 21,
8092 Zurich,
Switzerland.}

\email{\{karen.bodie, maximilian.brunner\}@mavt.ethz.ch}

\begin{abstract}
Omnidirectional micro aerial vehicles are a growing field of research, with demonstrated advantages for aerial interaction and uninhibited observation.
While systems with complete pose omnidirectionality and high hover efficiency have been developed independently, a robust system that combines the two has not been demonstrated to date.
This paper presents the design and optimal control of a novel omnidirectional vehicle that can exert a wrench in any orientation while maintaining efficient flight configurations.
The system design is motivated by the result of a morphology design optimization. A six degrees of freedom optimal controller is derived, with an actuator allocation approach that implements task prioritization, and is robust to singularities.
Flight experiments demonstrate and verify the system's capabilities.
\end{abstract}


\multimedia{A supporting video showcasing experiments can be accessed at \href{https://youtu.be/mBi9mOQaZzQ}{https://youtu.be/mBi9mOQaZzQ}}

\maketitle

\newcommand{\norm}[1]{\left\lVert#1\right\rVert}
\newcommand{\eye}[1]{\mathcal{I}_{#1}}

\renewcommand{\frame}[1]{\mathscr{F}_{#1}}
\newcommand{\world}{W}
\newcommand{\body}{B}
\newcommand{\rotor}[1]{{r_{#1}}}
\newcommand{\inertia}{\bm{J}}
\newcommand{\rot}[2]{\bm{R}_{{#1}{#2}}}
\newcommand{\pos}{\bm{p}}
\newcommand{\tmav}{TMAV}
\newcommand{\des}{d}

\section{Introduction}
\label{sec:intro}
Omnidirectional \acp{MAV} present a compelling solution for future applications of aerial robots. Full actuation allows for \emph{force-omnidirectionality} (force and torque tracking in six \ac{DOF}), decoupling the translational and rotational dynamics of the system, and permitting stable interaction with the environment. Such a system proposes significant functional advantages over the traditional underactuated \ac{MAV}, which provides only four controllable \acp{DOF} by nature of the aligned propeller axes. 

With the added criterion that force and torque in any direction must compensate for the system's mass, the result is complete \emph{pose-omnidirectionality}, where a system can achieve uninhibited aerial movement and robust tracking of six \ac{DOF} trajectories. This extension offers a unique advantage for aerial filming and 3D mapping, as well as configuration-based navigation in constrained environments.

A dominant struggle in the development of \acp{MAV} is reaching a compromise between performance and efficiency.
For robust aerial interaction and high control authority in omnidirectional flight, performance can be represented by the force and torque control volumes of a system.
For pose-omnidirectional platforms, the force envelope must exceed gravity in all directions with an additional buffer to maintain dynamic movement.
Countering this performance goal is the desire for high efficiency and longer flight times, which is compromised in systems that generate high internal forces (thrust forces which counteract each other), or add additional weight for actuation.

Within the past 5 years, substantial growth has occurred in the field of fully actuated omnidirectional \acp{MAV}, from the emergence of these systems to their application in realistic inspection scenarios.
We consider two dominant categories of platform actuation: fixed rotor, and tiltrotor platforms. The fixed rotor \ac{OMAV} offers a mechanically simple design to achieve full actuation, creating a thrust vector by varying propeller speeds of fixedly tilted rotors.
However, any interaction wrench exerted on the environment from a stable hovering pose creates a proportionately significant amount of internal force, which directly detracts from flight efficiency. As a result, orienting the propellers to prefer efficient hover flight and a higher payload directly reduces the capability to generate lateral force for interaction. Several fixed rotor platforms that achieve full pose omnidirectionality have been developed in recent years \citep{brescianini2016design, park2018odar, staub2018towards}, while other platforms achieve omnidirectional wrench generation from a defined hover orientation \citep{ryll2018truly,wopereis2018multimodal,ollero2018aeroarms}. Other concepts have extended the theory of fixed rotor platforms \citep{tognon2018omnidirectional}, as interest in these systems grows.

Tiltrotor platforms can individually tilt rotor groups with additional actuation, and can achieve optimal hover efficiency in the absence of external disturbances, when all propeller thrust vectors are aligned against gravity.
Tiltrotor systems have been demonstrated in the form of a quadrotor \citep{falconi2012dynamic, ryll2015novel} with limited roll and pitch,
and more recently in the form of a hexarotor \citep{kamel2018voliro}.
These platforms achieve force omnidirectionality with high hover efficiency in specific poses, at the cost of additional inertia and mechanical complexity, but force-omnidirectionality is marginal or impossible in some body orientations.
Another concept that reduces complexity by coupling the tilt axes to one or two motors has been evaluated in simulation \citep{ryll2016modeling, morbidi2018energy},
providing versatility in force generation and efficiency, but without pose-omnidirectionality.



When designing a system to target the application of aerial workers, versatility plays a large role. Navigation efficiency and payload capacity are often required, while new tasks of omnidirectional interaction demand high force capabilities in all directions. The tilt-rotor \ac{OMAV} provides a promising solution, having omnidirectional capabilities while maintaining the ability to revert to an efficient hover. While additional motors add complexity and weight, we can take advantage of the highly overactuated system to prioritize tasks in the allocation of actuator commands.

In this paper we present the design of an omnidirectional tiltrotor dodecacopter, a jerk-level \ac{LQRI} optimal controller for six \ac{DOF} control of a floating base system, and task prioritization in actuator allocation. Various experiments show controller performance, and demonstrate handling of singularities and individual tilt arm control.

\subsection{Contributions}
\begin{itemize}
\item Design of a novel \ac{OMAV} that achieves both force- and pose-omnidirectionality with highly dynamic capabilities, while maintaining high efficiency in hover. A tiltrotor design optimization tool is described and made available open source.
\item A full state optimal controller is designed in the form of a jerk-level \ac{LQRI}, which takes into account tilt-motor commands.
\item An allocation strategy is developed to prioritize tracking in 6 \ac{DOF}, while completing additional tasks in the null space of the overactuated system.
\item Experiments demonstrate the tracking performance of the \ac{LQRI} as compared to a benchmark \ac{PID} controller. Further tests show singularity handling and cable unwinding as secondary tasks while tracking a full pose trajectory.
\end{itemize}

\subsection{Outline}
This paper extends upon the work presented at the 2018 International Symposium on Experimental Robotics (ISER 2018) \citep{bodie2018towards} and \citep{bodie2019omnidirectional}, and is structured as follows. \Cref{sec:model} states assumptions and notation, and derives a general model of the tiltrotor \ac{OMAV}. \Cref{sec:design} describes the parametric design the system subject to a cost function, and compares the system's theoretical capabilities to other state-of-the-art concepts. \Cref{sec:control} details the \ac{LQRI} optimal control implementation, and describes the allocation of 6 \ac{DOF} wrench tracking commands, and additional secondary tasks, to 18 independent actuators. \Cref{sec:setup} describes the hardware of the prototype system and the experimental setup. In \cref{sec:experiments}, results from experimental flights are presented and discussed. Finally, \cref{sec:conclusion} offers concluding remarks and directions for future work.

\section{System Modeling}
\label{sec:model}
In this section we present definitions and notation used throughout the paper, define assumptions, and develop a generalized tiltrotor \ac{MAV} model. We further discuss singularity cases that arise in generalized tiltrotor systems.

\subsection{Definitions and notation}
\begin{figure}[h]
 \centering
\includegraphics{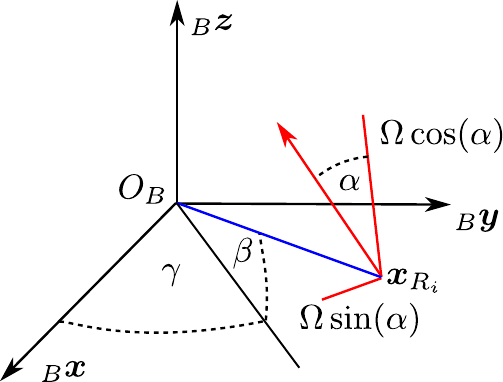}
 \caption{Illustration of arm angles.}
 \label{fig:arm_angle_definitions}
\end{figure}

In the present work, we consider a general rigid-body model for a tilt-rotor aerial vehicle.
Refer to \cref{tab:symbols} for definitions of common symbols used throughout the paper.
We use following conventions for coordinate frames:
\begin{itemize}
    \item Inertial world frame $\frame{\world}$
    \item Body frame $\frame{\body}$ attached to the \ac{COM} of the \ac{MAV} with origin $O_{\body}$
    \item Rotor frame $\frame{\rotor{i}}$
\end{itemize}
\Cref{fig:arm_angle_definitions} illustrates the frames and angles that we use when describing a tiltrotor \ac{MAV}.
The rotating axis of each rotor group, $x_{R_{i}}$, is defined as the axis of the arm coming out of the main body, and intersects the body origin, {$O_{\body}$}. This arm may be positioned at arbitrary angles $\gamma_{i}, \beta{i}$ which respectively define the rotation about the $z_{\body}$ axis, and inclination from the $z_{\body}$-plane. 

$\rot{\world}{\body}\in\mathrm{SO}(3)$ denotes a rotation matrix expressing the orientation of $\frame{\body}$ with respect to $\frame{\world}$.

$\f{\body}\vec{\omega}_{\world\body}\in\mathbb{R}^{3}$ denotes the angular velocity of $\frame{\body}$ with respect to $\frame{\world}$, expressed in $\frame{\body}$. We can then express the orientation kinematics of a frame $\frame{\body}$ as
\begin{equation}
    \dot{\bm{R}}_{\world\body}=\bm{R}_{\world\body}[\f{\body}\vec{\omega}_{\world\body}]_\times
\end{equation}
with $[\cdot]_\times\in\mathrm{SO}(3)$ as the skew symmetric matrix of the vector $\cdot\in\mathbb{R}^3$.
The angular acceleration of the body frame is written as $\f{\body}\vec{\psi}_{\world\body}$ and the angular jerk as $\f{\body}\vec{\zeta}_{\world\body}$.

Furthermore, we denote the position of the \ac{MAV} in the frame $\frame{\world}$ as $\f{\world}\vec{p}$. The body velocity, acceleration, and jerk are described as $\f{\world}\vec{v}, \f{\world}\vec{a}$, and $\f{\world}\vec{j}$, respectively.

The time derivative of a vector $\f{B}a$ in a rotating frame $\frame{B}$ is written as follows
\begin{equation}
    \frac{d}{dt}\left(\f{\body}\vec{a}\right)=\f{\body}\left(\dot{\vec{a}}\right)
\end{equation}
and can be expanded to
\begin{equation}\label{eq:kinematic_derivative}
    \f{\body}\left(\dot{\vec{a}}\right) = \f{\body}\dot{\vec{a}}+[\f{\body}\vec{\omega}_{\world\body}]_\times\f{\body}\vec{a}
\end{equation}

\begin{table*}[t]
    \centering
    \caption{Symbols and definitions for a tilt-rotor aerial vehicle.}
    \label{tab:symbols}
    \begin{tabularx}{\textwidth}{l l}
        \hline
        Symbol & Definition \\
        \hline
        $\frame{\world}: \{O_{\world},x_{\world},y_{\world},z_{\world}\}$ & inertial frame: origin and primary axes \\
        $\frame{\body}: \{O_{\body},x_{\body},y_{\body},z_{\body}\}$ & robot body-fixed frame: origin and primary axes \\
        $\frame{\rotor{i}}: \{O_{\rotor{i}},x_{\rotor{i}},y_{\rotor{i}},z_{\rotor{i}}\}$  & $i^{\text{th}}$ rotor unit frame: origin and primary axes \\
        $\rot{a}{b}$ & Orientation of $\frame{b}$ expressed in $\frame{a}$ \\
        $\bm{A},\bm{A}_{\alpha}$ & Static allocation matrix and instantaneous allocation matrix, respectively \\
        $m$ & mass\\
        $\inertia$ & inertia matrix\\
        $\vec{r}_{com}$ & center of mass offset, given in the body frame\\	
        $\vec{\alpha}$ & vector of tilt angles \\
        $\vec{\omega}$ & vector of angular velocities of the rotors \\
        $\vec{\Omega}$ & vector of squared rotor speeds \\
        $\tilde{\vec{\Omega}}$ & vector of lateral and vertical components of $\vec{\Omega}$ \\
        $\vec{f}$ & force vector \\
        $\vec{\tau}$ & torque vector \\
        $\f{\body}\vec{w}$ & total actuation wrench in the body frame \\
        $\f{\world}\vec{g} = [0 \; 0 \; g]^{\top}$ & gravity acceleration vector, $g = 9.81$ \si{\meter\second}$^{-2}$\\
        $\f{\world}\vec{p}$ & position \\  		
        $\f{\world}\vec{v}$ & linear velocity \\ 		
        $\f{\world}\vec{a}$ & linear acceleration \\
        $\f{\world}\vec{j}$ & linear jerk \\
        $\f{\body}\vec{\omega}_{\world\body}$ & angular velocity\\
        $\f{\body}\vec{\psi}_{\world\body}$ & angular acceleration\\
        $\f{\body}\vec{\zeta}_{\world\body}$ & angular jerk\\
        \hline
    \end{tabularx}
\end{table*}

\subsection{Assumptions}
The following assumptions are adopted to simplify the model:
\begin{itemize}
 \item The entire platform can be modeled as a rigid body.
 \item Thrust and drag torques are proportional to the square of the rotor’s angular speed, and rotors are able to achieve desired speeds $\omega_i$ with negligible transients.
 \item The primary axes of the system correspond with the principal axes of inertia. Products of inertia are considered negligible.
 \item The dynamics of tilt motors are independent of the rotational speed of rotors.
 \item The tilting axis of each propeller group intersects the body origin.
 \item The thrust and drag torques produced by each rotor are independent, i.e. there is no airflow interference.
\end{itemize}

\subsection{Rigid body model}
The system dynamics are derived by the Newton-Euler approach under the stated assumptions,
resulting in the following equations of motion expressed in the body-fixed frame:

\begin{multline}\label{eq:EOM}
 \begin{bmatrix}
  m \eye{3} & 0          \\
  0            & \bm{J}_{\body}
 \end{bmatrix}
 \begin{bmatrix}
  \f{\body}\dot{\bm{v}}    \\
  \f{\body}\dot{\bm{\omega}}_{\world\body}
 \end{bmatrix}
 =-
 \begin{bmatrix}
  \left[\f{\body}\bm{\omega}_{\world\body}\right]_\times{} m \f{\body}{\bm{v}}_{\body}      \\
  \left[\f{\body}\bm{\omega}_{\world\body}\right]_\times \inertia \f{\body}\vec{\omega}_{\world\body}
 \end{bmatrix}\\
 +\begin{bmatrix}\f{\body}\vec{f} \\ \f{\body}\vec{\tau}\end{bmatrix} + 
 \begin{bmatrix}
 m\f{\body}\vec{g}\\ 0
 \end{bmatrix}
 \end{multline}
 
where $m$ and $\inertia$ are the mass and inertia of the \ac{MAV}, $\begin{bmatrix}\f{\body}\vec{f}^T & \f{\body}\vec{\tau}^T\end{bmatrix}^T$ are forces and torques resulting from the actuation, and $m\f{\body}\vec{g}$ is the force resulting from gravity.

\subsection{Aerodynamic force model}
For any geometry of a tiltrotor \ac{MAV} we can find a relation between the individual rotor forces $\vec{f}_{r,i}$ and torques $\vec{\tau}_{r,i}$ and the total body force and torque vector $\f{\body}\vec{w}$ that is generated at its center of mass.

As previously presented by \cite{kamel2018voliro}, we assume that the force and torque produced by a rotor can be described as
\begin{subequations}
\begin{align}
    f_i &= c_F\omega_i^2\\
    \tau_i &= c_M\omega_i^2
\end{align}
\end{subequations}
Since both the force and torque are directly proportional to the squared rotor speed, we define the vector of squared rotor speeds:
\begin{equation}
    \bm{\Omega} = \begin{bmatrix}\Omega_{1} \\ \vdots \\ \Omega_{n} \end{bmatrix} = \begin{bmatrix}\omega_{1}^2 \\ \vdots \\ \omega_{n}^2 \end{bmatrix}
\end{equation}
Considering a tiltrotor model, $\alpha$ represents the active rotational joint angle about an arm axis. We define the \emph{instantaneous allocation matrix} $\bm{A}_\alpha$ that is a nonlinear function of the current tilt angles $\vec{\alpha}=\begin{bmatrix}\alpha_1,&\dots,&\alpha_n\end{bmatrix}$ with $n$ being the number of tilt arms.
Using a matrix multiplication we can relate the body force and torque directly with the squared rotor speeds:
\begin{equation}
    \f{\body}\vec{w}=\begin{bmatrix}\f{\body}\vec{f} \\ \f{\body}\vec{\tau}\end{bmatrix}
 = \bm{A}_{\alpha} \vec{\Omega}
\end{equation}
For convenience we can also write this equation with a static allocation matrix $\bm{A}$ that is independent of the varying tilt angles and that only depends on the geometry of the \ac{MAV}:
\begin{equation}
 \f{\body}\vec{w} = \bm{A} \tilde{\vec{\Omega}} \qquad \bm{A}\in\mathbb{R}^{6\times n},\tilde{\vec{\Omega}}\in\mathbb{R}^{2n}
 \label{eq:static_allocation}
\end{equation}
\begin{equation}
 \begin{matrix}
  \tilde{\vec{\Omega}} =
  \begin{bmatrix}
  \sin(\alpha_i) \Omega_{i}  \\
  \cos(\alpha_i) \Omega_{i}  \\
  \vdots \\
  \sin(\alpha_n) \Omega_{n} \\
  \cos(\alpha_n) \Omega_{n}
  \end{bmatrix}
  &
  \forall i \in \{1...n\}
 \end{matrix}
\end{equation}
The elements of the vector $\tilde{\vec{\Omega}}$ correspond to the lateral and vertical components of each squared rotor speed in the respective rotor unit frame.

Refer to \cref{sec:appendix_allocation} for a general formulation of the static allocation matrix $\bm{A}$.

\subsection{Tilt motor model}
As proposed by \cite{kamel2018voliro} we model the tilt angle dynamics using a first order damped system, i.e.
\begin{equation}
    \dot\alpha_i = \frac{1}{\tau_\alpha}\left(\alpha_{i,ref}-\alpha_i\right)
\end{equation}
with $\alpha_{i,ref}$ being the reference, $\alpha_i$ the actual tilt angle of arm $i$, and $\tau_\alpha$ the time constant of the tilting motion.
This accounts for dynamic effects as well as physical limitations of the servo motors that are otherwise unmodeled.

\subsection{Singularity cases for tiltrotor MAVs}
As previously identified by \cite{bodie2018towards}, an \ac{OMAV} can encounter two types of singularity cases that result from the allocation model.

The first type occurs when the tilt angles $\alpha_i$ are aligned in a way that leads to a rank reduction of the instantaneous allocation matrix $\bm{A}_\alpha$. This corresponds to a state in which instantaneous controllability of select forces and torques is lost due to tilt motor delay. We refer to this type of singularity as a \emph{rank reduction singularity}. It has been studied thoroughly by \cite{morbidi2018energy}.

The second type, which we refer to as a \emph{kinematic singularity}, occurs when rotor thrusts cannot contribute to the desired body wrench $\vec w_{\body}$, which leads to tilt angles not being uniquely defined. This condition has been described by \cite{bodie2018towards}.

\section{Tiltrotor morphology design}
\label{sec:design}

The tiltrotor \ac{MAV} offers compelling advantages toward the goal of versatility in force exertion and hover efficiency. However, despite these advantages, tiltrotor systems also come with drawbacks such as additional actuation mass and complexity, limited rotation due to possible arm cable windup, and the presence of singularity cases that are not encountered in a fixed rotor system. Therefore, morphology design is important to ensure that the resulting platform meets performance requirements. This section describes the morphology design approach, compares this resulting morphology to other state of the art omnidirectional systems, and justifies the choice of a tiltrotor \ac{OMAV}.

\begin{figure}[htp]
 \centering
\includegraphics[width=\linewidth]{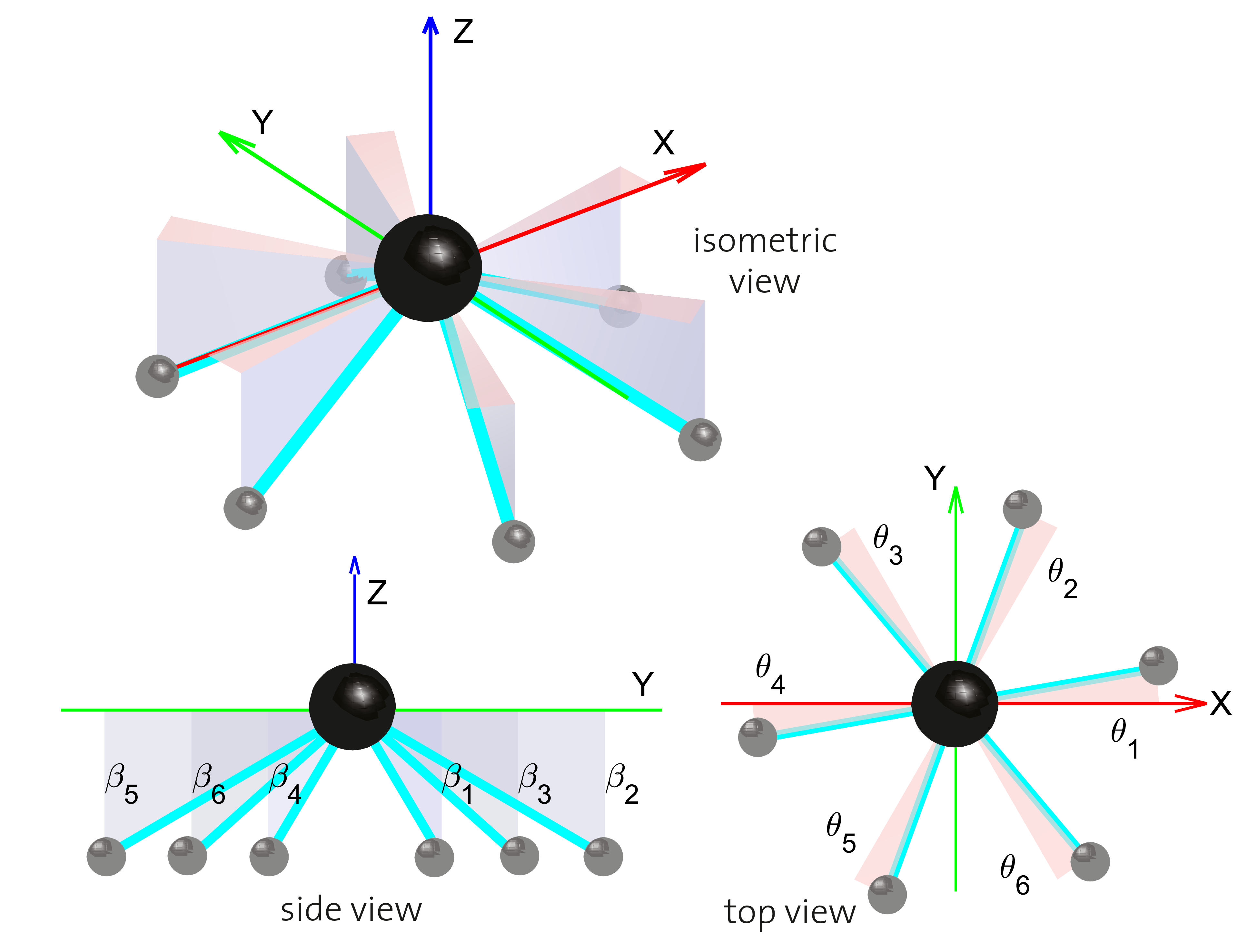}
 \caption{Definitions of angles $\theta$ and $\beta$ for the tilt-rotor morphology optimization.}
 \label{fig:design_images}
\end{figure}

\subsection{Evaluation metrics}
We aim to achieve the following design goals:
\begin{itemize}
    \item Fully actuated system in any hover orientation.
    \item High force and torque capabilities in all directions.
    \item High efficiency hover in at least one orientation.
\end{itemize}

To evaluate the dynamic capabilities of force- and pose-omnidirectionality, as well as hover efficiency, we define the following metrics: 

\subsubsection{Force and torque envelopes }
Volumes corresponding to the maximum reachable forces ($f_{vol}$) and torques ($\tau_{vol}$) of the system expressed in $\frame{\body}$ can be computed by feeding commands forward through the actuator allocation matrix (see \cref{eq:static_allocation}). We refer to these volumes as \emph{force} and \emph{torque envelopes}.
Maximum, minimum, and mean values for the envelopes are calculated ($f_{max},f_{min},f_{mean}$ and $\tau_{max},\tau_{min},\tau_{mean}$) and used to evaluate dynamic capabilities. The force envelope is computed in the absence of torque, and the torque envelope is computed in the presence of a static hover force. Torque envelopes presented here include a hover force aligned with $z_{\body}$, though other directions have been evaluated to verify full pose omnidirectionality in all hover conditions for optimized tiltrotor platforms.

\subsubsection{Force efficiency index, $\eta_{f}$ }
As an efficiency metric of omnidirectional force exertion, we use a \textit{force efficiency index} $\eta_{f} \in [0,1]$ as originally defined by \cite{ryll2016modeling}. The index represents the ratio of the desired body force magnitude to the sum of individual rotor group thrust magnitudes, as expressed in \cref{eq:force_index}. 

\begin{equation}
    \eta_{f} = \frac{\norm{\f{\body}\bm{f}_d}}{\sum_{i=1}^n{\f{\body}f_{\rotor{i}}}}
    \label{eq:force_index}
\end{equation}

When $\eta_{f} = 1$, no internal forces are present, and all acting forces are aligned with the desired force vector $\bm{f}_{d}$. While these internal forces should be reduced for efficient flight, they also allow for instantaneous disturbance rejection, since thrust vectoring can be achieved by changing only rotor speeds.

A \emph{torque efficiency index} can be computed similarly in \cref{eq:torque_index} to evaluate the efficiency of maximum torque commands, considering the body moment due to propeller forces acting at distance $l$ from $O_{\body}$. Torque contributions due to rotor drag are assumed to be negligible, since they are a significantly smaller multiple of the thrust force.

\begin{equation}
    \eta_{\tau} = \frac{\norm{\f{\body}\bm{\tau}_d}}{l\sum_{i=1}^n{\f{\body}f_{\rotor{i}}}}
    \label{eq:torque_index}
\end{equation}

\subsubsection{System Inertia }
The inertial properties of the platform translate generalized force to acceleration, and therefore shape the agility of the flying system. Reducing inertia increases agility, and can be achieved by reducing the total mass $m$, and bringing the mass closer the the body origin $O_{\body}$. Dynamic tiltrotor inertial effects are neglected for this analysis. 
The mass and inertia used at each optimization step are parametrically computed based on a simplified system geometry and realistic component masses, as derived in \cref{app:inertia}.

\subsection{Optimization}
We present a tiltrotor morphology optimization tool, made available open source\endnote{Tiltrotor morphology optimization tool is available open source at \href{https://github.com/ethz-asl/tiltrotor_morphology_optimization}{github.com/ethz-asl/tiltrotor\_morphology\_optimization}}. 
This tool performs parametric optimization of a generalized tiltrotor model as shown in \cref{fig:design_images}, subject to a cost function of force and torque exertion, agility and efficiency metrics. We define fixed angles $\bm{\theta},\bm{\beta} \in \mathbb{R}^{n}$ as $z_{\body}$- and $y_{\body}$-axis angular deviations from a standard multicopter morphology with arms evenly distributed in the $z_{\body}$-plane. Rotor groups can tilt actively about $x_{\rotor{i}}$, and are controlled independently. For the purposes of this paper, we only optimize over $\bm{\theta}$ and $\bm{\beta}$. Although the tool allows for additional optimization over number of arms $n$ and arm length $l$, we set these parameters in advance to $6$ and \SI{0.3}{\meter} respectively.

We consider two cost functions to evaluate an omnidirectional tiltrotor system: one that prefers unidirectional flight efficiency while requiring omnidirectional flight, and another that maximizes omnidirectional force and torque capabilities. The first cost function is defined as a maximization of the force envelope in one direction (here, unit vector $\hat{\bm{z}}_{\body}$), while requiring a minimum force in all directions, expressed as

\begin{equation}
\begin{aligned}
& \min_{\bm{\theta},\bm{\beta}} (-f_{max} \hat{\bm{z}}_{\body}) \\
\text{s.t.} &  \\
   & f_{min} > mg \\
    &-\frac{\pi}{2} < \theta_i < \frac{\pi}{2} \\
    &-\frac{\pi}{2} < \beta_i < \frac{\pi}{2} \\
    &\omega_{min} < \omega_i < \omega_{max},  \quad \forall i \in \{1\dots n\}
\end{aligned}
\label{eq:cost1}
\end{equation}

The second cost function seeks to maximize omnidirectional capability by maximizing the minimum force and torque envelope values in all body directions, i.e. for all surface point directions on a unit sphere. This result will guarantee omnidirectional hover if the system's parameters provide a sufficient thrust to weight ratio.

\begin{equation}
\begin{aligned}
&\min_{\theta,\beta} (-f_{min},-\tau_{min}) \\
\text{s.t.} & \\
& f_{min} > mg \\
& -\frac{\pi}{2} < \theta_i < \frac{\pi}{2} \\
& -\frac{\pi}{2} < \beta_i < \frac{\pi}{2} \\
& \omega_{min} < \omega_i < \omega_{max}, \quad \forall i \in \{1\dots n\}
\end{aligned}
\label{eq:cost2}
\end{equation}

\begin{table}[t]
\begin{threeparttable}
\caption{Optimized morphologies and simplified metrics}
\label{tab:optimization_results}
\centering
\def\arraystretch{1.2}
\begin{tabularx}{\columnwidth}{l c}
\hline
property & cost function 1, \cref{eq:cost1}\\
\hline
$\bm{\beta}$ [deg] 
& $[0,0,0,0,0,0]$ \\
$\bm{\theta}$ [deg] 
& $[0,0,0,0,0,0]$ \\
mass [kg] 
& $4.0$ \\
inertia$^{*}$ [kgm$^2$] 
& $\{0.0725, 0.0725, 0.1439\}$ \\
\{$f_{min}$, $f_{max}$\}[N]
& $\{72.4, 144.7\}$ \\
$f_{vol}$[N$^3$] 
& $3.5 \cdot 10^6$ \\
\{$\tau_{min}$, $\tau_{max}$\}[Nm]
& $\{26.5, 68.4\}$ \\
$\tau_{vol}$[Nm$^3$] 
& $2.1 \cdot 10^5$ \\
$\eta_{f}$ at hover \{min, max\} 
& $\{0.75, 1.0\}$ \\
\hline
\hline
property & cost function 2, \cref{eq:cost2}\\
\hline
$\bm{\beta}$ [deg] 
& $35.26 \cdot [1, -1, 1 -1, 1 -1]$ \\
$\bm{\theta}$ [deg] 
& $[0,0,0,0,0,0]$ \\
mass [kg] 
& $4.0$ \\
inertia$^{*}$ [kgm$^2$] 
& $\{0.0959, 0.0959, 0.0972\}$ \\
\{$f_{min}$, $f_{max}$\}[N]
& $\{96.5, 118.2\}$ \\
$f_{vol}$[N$^3$] 
& $4.2 \cdot 10^6$ \\
\{$\tau_{min}$, $\tau_{max}$\}[Nm]
& $\{28.7, 44.0\}$ \\
$\tau_{vol}$[Nm$^3$] 
& $1.7 \cdot 10^5$ \\
$\eta_{f}$ at hover \{min, max\} 
& $\{0.82, 1.0\}$ \\
\hline
\end{tabularx}
\begin{tablenotes}
  \item[] $^{*}$Primary components of inertia are presented, products of inertia are assumed negligible.
\end{tablenotes}
\end{threeparttable}
\end{table}

\begin{figure*}[tp]
 \centering
\includegraphics[width=\textwidth]{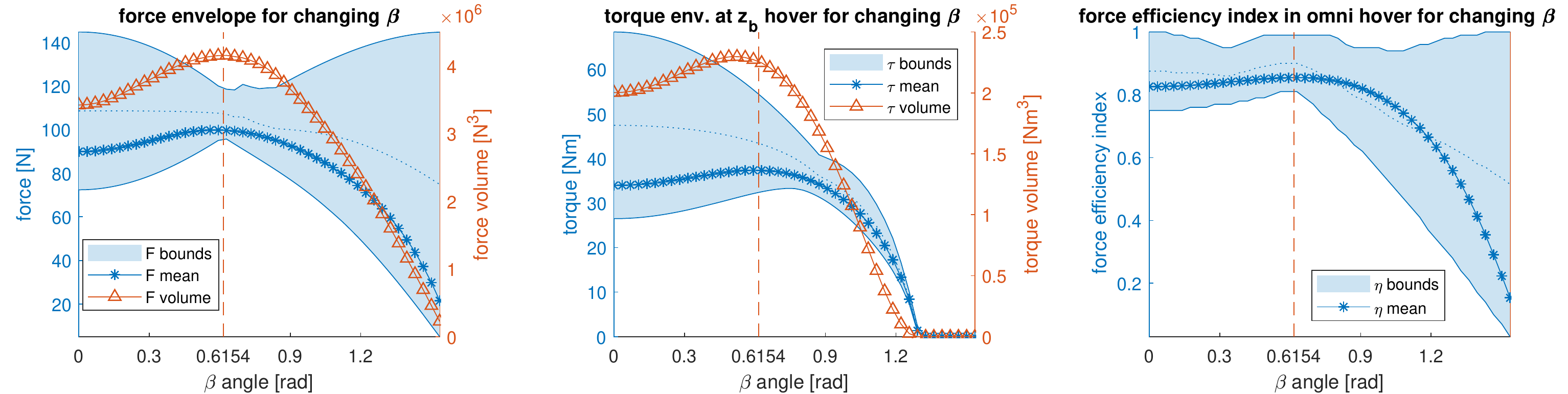}
 \caption{Variation of force and torque envelope metrics (max and min bounds, mean and volume) with changing $\beta$, the most efficient hover solution is at $\beta = 0$ and the maximized force/torque envelope is at $\beta = 0.6154$ rad ($35.26$ deg).}
 \label{fig:design_trends}
\end{figure*}

The resulting optimized morphologies and metrics are presented in \cref{tab:optimization_results}. 
The first optimization function results in a tiltrotor that takes a standard hexacopter morphology. Force in a single direction is maximized along $z_{\body}$, and system properties are sufficient to ensure hover in any orientation.
The second optimization has multiple solutions of the same result result metrics, placing rotor groups $O_{\rotor{i}}$ at the vertices of an arbitrarily oriented octahedron. We list the $\bm{\beta}$ values when $\bm{\theta}$ are $0$. Results show that both systems can sustain omnidirectional hover with additional force capability, and have omnidirectional force and torque envelopes, as seen by the $f_{min}, \tau_{min}$. We further evaluate the evolution of performance metrics between solutions, with $\bm{\theta}$ values fixed at $0$, and $\bm{\beta} = \beta \cdot [1,-1,1,-1,1,-1]$ for changing $\beta$. Results are shown in \cref{fig:design_trends}, where we see a clear maximization of the minimum reachable force in the left plot. Considering the goal of versatility for efficient flight with omnidirectional capabilities, the $\bm{\beta} = 0$ solution presents high reachable forces with a maximum efficiency index, and sufficiently large minimum reachable forces and torques for agile flight in 6 \ac{DOF}. 

\subsection{Comparison}
We compare capabilities of the two optimized tiltrotor \ac{OMAV} results to other state-of-the-art \acp{OMAV}. Platforms selected for comparison are the fixed rotor \textit{tilt-hex} realized for aerial interaction applications \citep{ryll20196d} and the fixed rotor omnidirectional platforms derived and realized by \cite{brescianini2016design, park2016design}.

\begin{figure*}[tp]
 \centering
\includegraphics[width=\textwidth]{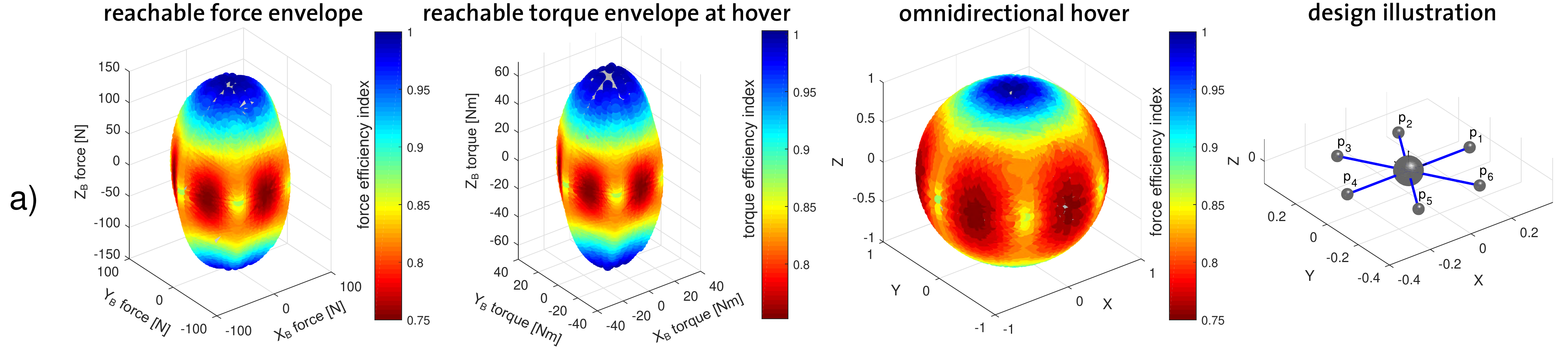}
\includegraphics[width=\textwidth]{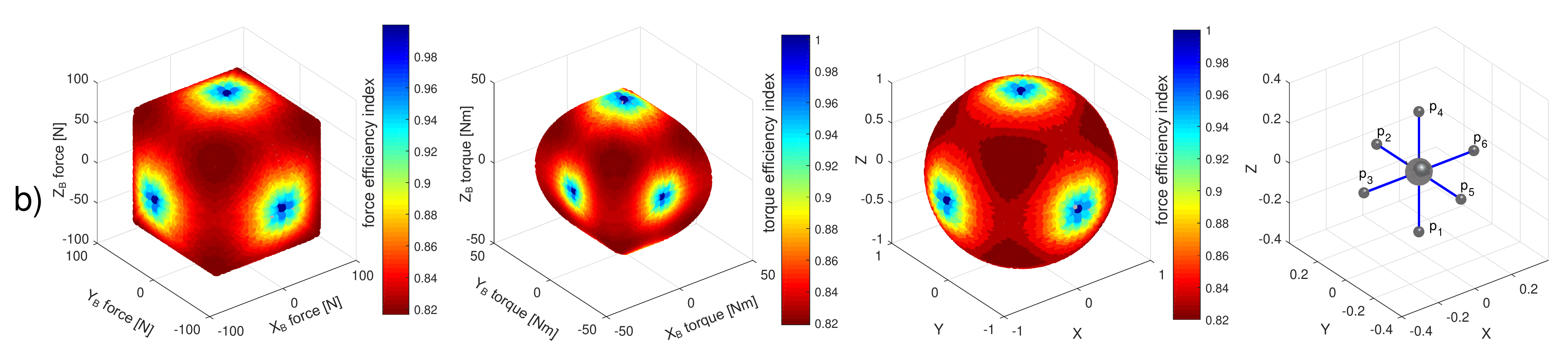}
\includegraphics[width=\textwidth]{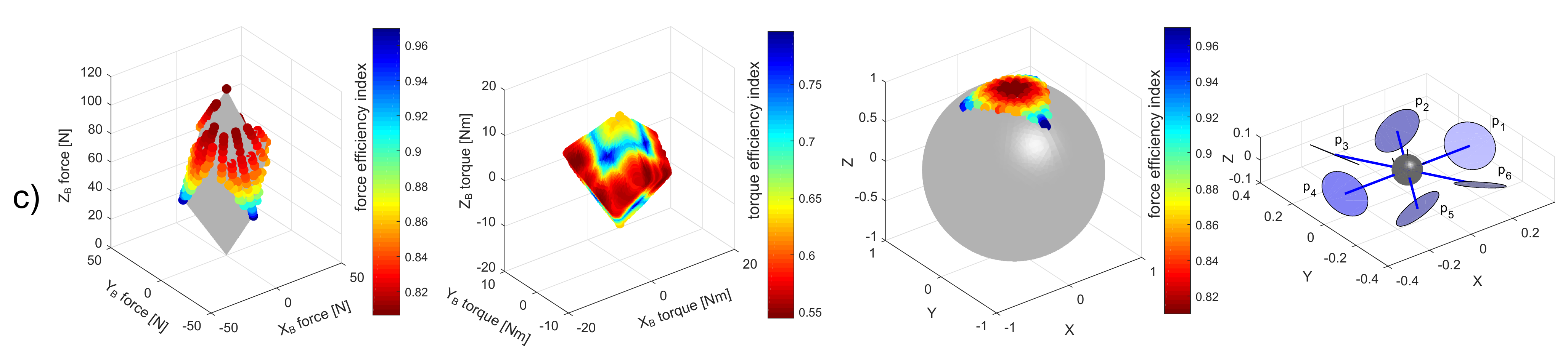}
\includegraphics[width=\textwidth]{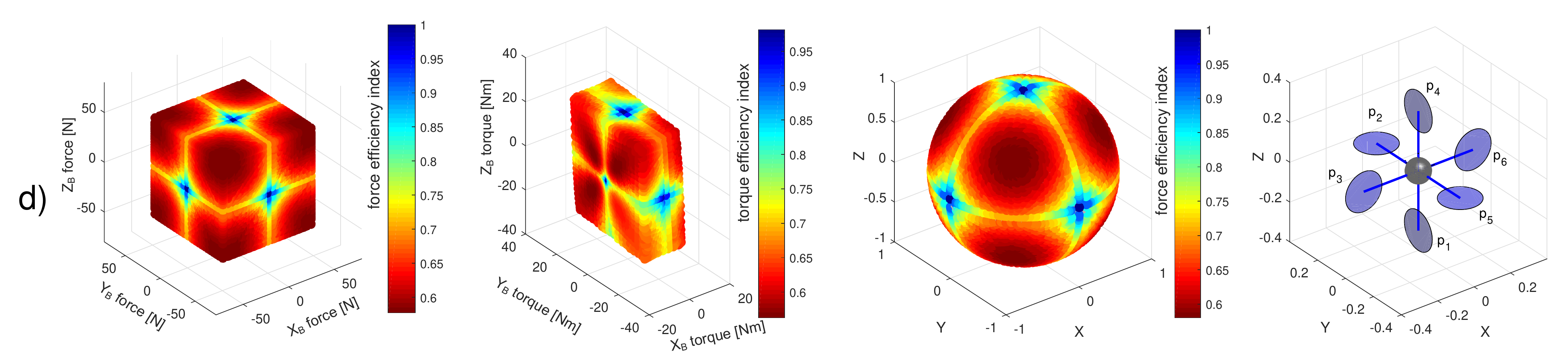}
 \caption{Reachable force envelope, reachable torque envelope at $z_{\body}$ hover, omnidirectional hover directions colored by force efficiency index, all expressed in $\frame{\body}$ and a morphology design illustration are shown for (a) a tiltrotor optimized for unidirectional efficiency, (b) a tiltrotor that maximizes $f_{min}$ and $\tau_{min}$, (c) a fixed rotor design from \cite{ryll20196d}, and (d) a fixed rotor design from \cite{brescianini2016design, park2016design}.}
 \label{fig:design_comparison}
\end{figure*}

For all platforms compared in \cref{fig:design_comparison}, we show result metrics in \cref{tab:comparison} relative to the first tiltrotor platform. For intuitive comparison of the second optimized tiltrotor result, the configuration is rotated such that axes align with the fixed rotor platform in \cref{fig:design_comparison} d).

\begin{table*}[t]
\begin{threeparttable}
\caption{Comparison of relative omnidirectional system properties corresponding to morphologies in \cref{fig:design_comparison}}
\label{tab:comparison}
\centering
\def\arraystretch{1.2}
\begin{tabularx}{\textwidth}{l c c c c}
\hline
property$^{*}$ & tiltrotor a) & tiltrotor b) & fixed rotor c) & fixed rotor d) \\
\hline
inertia$^{**}$ \{$m$, $I_{xx}$, $I_{yy}$, $I_{zz}$\} & reference & \{1.0, 1.33, 1.33, 0.67\} & \{0.77, 0.99, 0.99, 0.99\} & \{0.77, 1.32, 1.32, 0.67\} \\
force \{min, max, vol\} & reference & \{1.33, 0.81, 1.20\} & \{0, 0.81, 0.01\} & \{0.67, 0.58, 0.26\} \\
torque \{min, max, vol\} & reference & \{1.14, 0.66, 0.95\} & \{0.21, 0.17, 0.01\} & \{0.31, 0.53, 0.17\} \\
$\eta_f$ at hover \{min, max\} & reference & \{1.09, 1.0\} & \{1.08, 0.97\} & \{0.7, 1.0\} \\
\hline
\end{tabularx}
\begin{tablenotes}
  \item[] $^{*}$All values are expressed relative to the first tiltrotor system.
  \item[] $^{**}$Primary components of inertia are presented, products of inertia are assumed negligible.
\end{tablenotes}
\end{threeparttable}
\end{table*}

 Force and torque envelopes are colored with the efficiency index of the maximum achievable values in each direction. The third column shows the force efficiency index for each achievable hover direction, plotted on a unit sphere. As expected, the reachable force and torque envelopes for the two tiltrotor systems are much larger than their fixed rotor counterparts. Designs a) and c) achieve high forces in level hover, while designs b) and d) show a more uniform distribution of omnidirectional force. Due to its tilting rotors, design a) still has better omnidirectional force and torque characteristics than the fixed rotor design d) which is optimized for omnidirectional force.

The tiltrotor implementation of a standard hexacopter design promises a versatile and capable morphology solution. Additional weight and complexity can be justified by significantly improved performance metrics. Independent tilting of each rotor group results in overactuation: 12 inputs to control 6 \ac{DOF}. The controller can act in the null space of the allocation to assign secondary tasks.
We can further justify the tilt-rotor version of a standard hexacopter morphology for perception applications, where the dual unobstructed hemispheres of the $z_{\body}$ plane allow for a large field of view.

\begin{figure}[tp]
\centering
\includegraphics[width=\columnwidth]{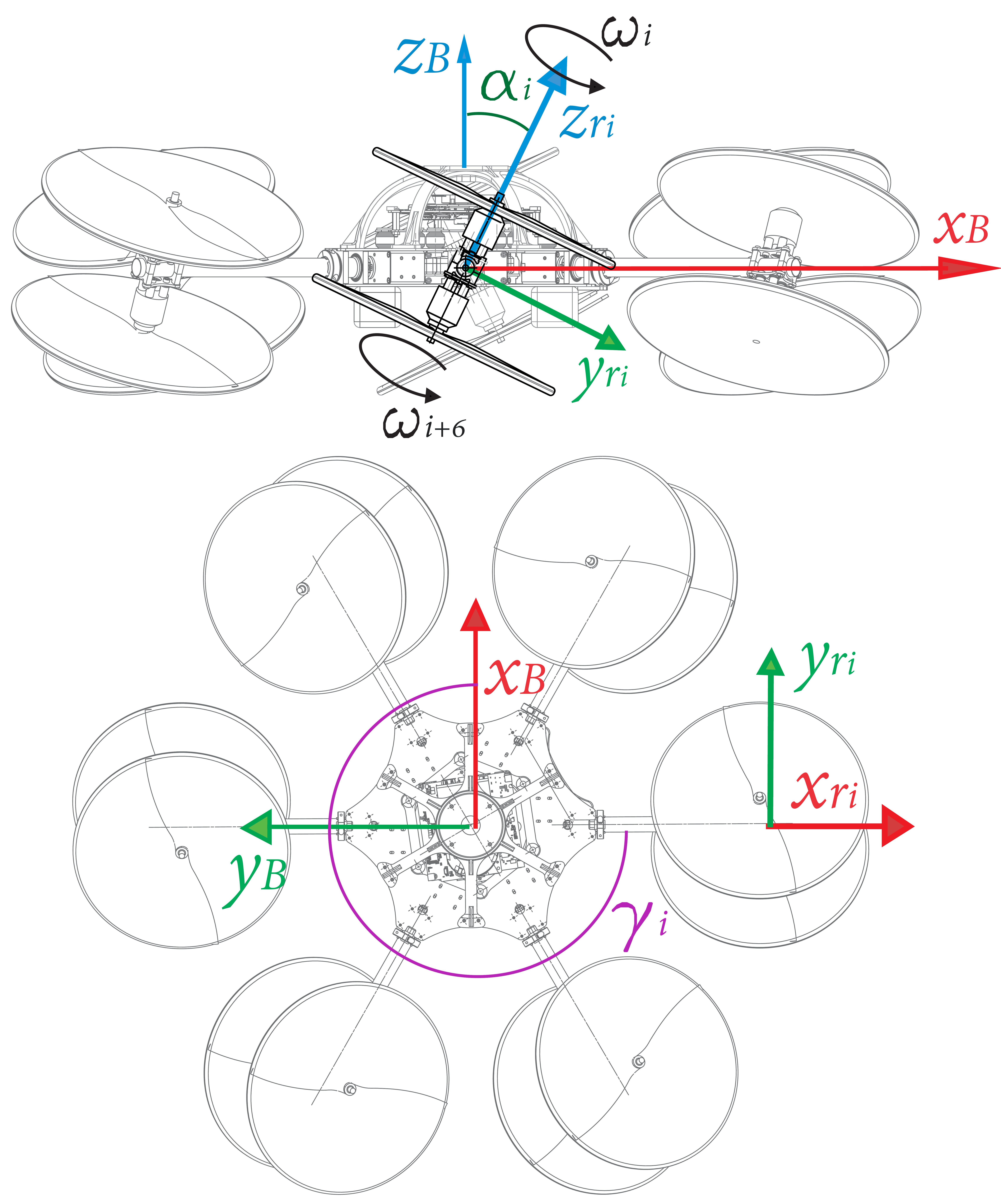}
 \caption{Coordinate frames and variables for $i^{\text{th}}$ rotor group $\rotor{i}$ and body $\body$. Principal axes are shown, as well as rotor group tilt angle $\alpha_{i}$, rotor angular velocity $\omega_{i}$, and arm spacing angle $\gamma_{i}$.}
 \label{fig:description}
\end{figure}

As previously discussed, dynamic inertial effects of tilting rotor groups were not taken into account for the optimization. As such, we choose to augment the system with double propeller groups to increase total thrust and to balance rotational inertia about the tilting axis. Counter-rotating propellers in each rotor group further reduce gyroscopic effects. Upper rotor directions alternate with adjacent arms to reduce the unmodelled effects of dual rotor groups.

Coordinate systems for the platform are described in \cref{fig:description}, with the body and rotor group frames $\frame{\body},\frame{\rotor{i}}$, and definitions of the fixed arm spacing angles $\bm{\gamma} = \frac{\pi}{6}\cdot[1,3,5,7,9,11]$ and tilting angles $\alpha_i$. 
Individual rotor angular velocities $\omega_{i}$ for $i \in (1,6)$ represent upper rotors and $i \in (7,12)$ represent lower rotors.

\section{Control}
\label{sec:control}
\subsection{Overview}
The control problem consists of tracking a 6 \ac{DOF} reference trajectory given by $\left(\vec p_{\des}(t),\bm{R}_{\des}(t)\right)\in\mathbb{R}^3\times\mathrm{SO}(3)$.
Previous works \cite{bodie2019omnidirectional,kamel2018voliro,ryll2015novel} exploit the rigid body model from \cref{eq:EOM} in combination with the aerodynamic force model from \cref{eq:static_allocation} to compute actuator commands based on a thrust vectoring control. Force and torque commands are computed from position and attitude errors respectively and actuator commands can be resolved using the Moore-Penrose pseudo-inverse of the static allocation matrix.
\begin{equation}
    \label{eq:alloc_inv}
    \tilde{\vec{\Omega}} = 
    \bm{A}^\dagger
    \begin{bmatrix}
    \f{\body}\vec{f}\\
    \f{\body}\vec{\tau}\\
    \end{bmatrix}
    \qquad
    \bm{A}^\dagger\in\mathbb{R}^{n\times6}
\end{equation}

While this method shows good performance in most flight scenarios, it exhibits some disadvantages:
\begin{enumerate}
    \item Singularities in the allocation matrix need to be avoided or handled explicitly
    \item Rotor and tilt arm dynamics are not considered by the allocation
\end{enumerate}

While controllers for overactuated \acp{MAV} usually operate on the acceleration level, we present two controllers that generate linear and angular jerk commands rather than acceleration commands. Specifically, we present
\begin{itemize}
    \item an \ac{LQRI} controller based on a rigid body model assumption, and
    \item a \ac{PID} controller that generates acceleration commands with subsequent numerical differentiation to obtain jerk.
\end{itemize}

The main motivation to use jerk rather than acceleration commands is that this allows us to access the tilt angle dynamics of the rotating arms. Therefore we also present a new method of allocation, based on optimizing over the differential actuation commands $\left(\dot{\bm{\alpha}},\dot{\bm{\omega}}\right)$ rather than the absolute actuation commands. We will refer to this as the \emph{differential allocation}.
This approach accounts for limited changes in tilt angles and rotor speeds, and can achieve almost arbitrary sets of tilt angles.

\begin{figure*}[t]
 \centering
\includegraphics[width=\textwidth]{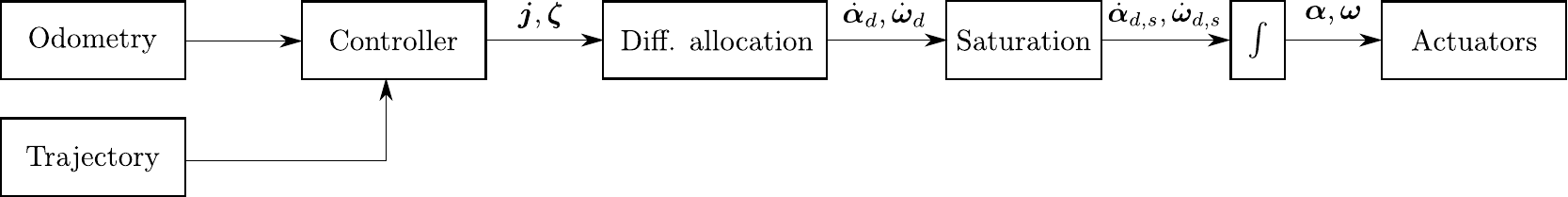}
 \caption{Control block diagram.}
 \label{fig:complete_control_diagram}
\end{figure*}

\Cref{fig:complete_control_diagram} gives a high-level overview of the controller block diagram. \ac{LQRI} or \ac{PID} can be used interchangeably in the control block to compute jerk commands that are resolved to actuator commands by the differential allocation.

In this section, we will use the following notation:
$\vec{u}=\begin{bmatrix}\f{B}\vec{j}^T&\f{B}\vec{\zeta}^T\end{bmatrix}^T$ denotes the commanded body jerk which corresponds to the input to the differential allocation. $\bar{\vec{u}}$ denotes the input after feedback linearization, and $\tilde{\vec{u}}=\begin{bmatrix}\dot{\vec{\omega}}^T & \dot{\vec{\alpha}}^T\end{bmatrix}^T$ is the set of differential actuator commands.

\subsection{LQRI controller}\label{sec:lqri}
We use an LQRI controller to optimize the system dynamics according to the following infinite time optimization problem:
\begin{equation}\label{eq:lqri}
\begin{aligned}
    \min_{\bar{\vec{u}}(t)}\ &\int_0^\infty \vec{e}(t)^T\bm{Q}\vec{e}(t) + \bar{\vec{u}}(t)^T\bm{R}\bar{\vec{u}}(t) dt\\
    &\text{s.t.}\quad \dot{\vec{e}}(t) = f( \vec{e}(t), \bar{\vec{u}}(t))
\end{aligned}\\
\begin{matrix}
    \vec{e}\in\mathbb{R}^{n}\\
    \bar{\vec{u}}\in\mathbb{R}^{m}\\
    \bm{Q}\in\mathbb{R}^{n\times n}\\
    \bm{R}\in\mathbb{R}^{m\times m}
    \end{matrix}
\end{equation}
where $\vec{e}$ is the error vector that is to be minimized, $\bar{\vec{u}}$ is the control input to the system, $f(\cdot)$ is the function describing the state evolution, and $\bm{Q}, \bm{R}$ are weighting matrices for the error and control input.

In the following sections we present the underlying system model and the inputs that are used for optimization.

\subsubsection{System jerk dynamics}
The dynamics of the model are obtained by differentiating the dynamic equations of a rigid body model:
\begin{align}
    \frac{d}{dt}\left( m\f{\body}\vec{a}\right) &= \frac{d}{dt}\left(\f{\body}\vec{f} + \rot{\body}{\world}\f{\world}\vec{g}\right)\nonumber\\
    m\f{\body}(\dot{\vec{a}}) &= \f{\body}(\dot{\vec{f}}) -[\f{\body}\vec{\omega}_{\world\body}]_\times \rot{\body}{\world}\f{\world}\vec{g}
\end{align}
where $\f{\body}(\dot{\vec{a}}=\f{\body}\vec{j}$ is the jerk in the body frame.
Similarly we differentiate the angular acceleration:
\begin{align}
    \label{eq:eom_ang_dt}
    \begin{split}
    \frac{d}{dt}\left( \inertia\f{\body}\vec{\psi}_{\world\body}+[\f{\body}\vec{\omega}_{\world\body}]_\times\inertia\f{\body}\vec{\omega}_{\world\body} \right) =\\
    \frac{d}{dt}\left( \vec{\tau}_{\body} -[\vec{r}_{com}]_\times\f{\body}\vec{f} \right)
    \end{split}
\end{align}
and obtain the following angular jerk equation of motion:
\begin{multline}\label{eq:ang_jerk}
    \inertia\f{\body}\dot{\vec{\psi}}_{\world\body} 
    + 2[\f{\body}\vec{\omega}_{\world\body}]_\times \inertia\f{\body}\vec{\psi}_{\world\body}\\
    + [\f{\body}\vec{\psi}_{\world\body}]_\times\inertia\f{\body}\vec{\omega}_{\world\body}
    +[\f{\body}\vec{\omega}_{\world\body}]_\times^2\inertia\f{\body}\vec{\omega}_{\world\body} \\
    =  \f{\body}(\dot{\vec{\tau}}) - \left[[\f{\body}\vec{\omega}_{\world\body}]_\times\vec{r}_{com}\right]_\times\f{\body}\vec{f}\\ -[\vec{r}_{com}]_\times\f{\body}(\dot{\vec{f}})
\end{multline}
Using the kinematics from \cref{eq:kinematic_derivative} we can also write the angular jerk explicitly as
\begin{equation}
    \f{\body}\vec{\zeta}_{\world\body} = \f{\body}(\dot{\vec{\psi}}_{\world\body}) = \f{\body}\dot{\vec{\psi}}_{\world\body} + [\f{\body}\vec{\omega}_{\world\body}]_{\times}\f{\body}\vec{\psi}_{\world\body}
\end{equation}

\subsubsection{Error vector}
We define the error state vector as a concatenation of linear and angular differences between the state and the reference. Reference variables are denoted with a subscript $(\cdot)_\des$.
\begin{equation}\label{eq:error_terms}
    \vec{e} = 
    \begin{bmatrix}
    \f{\world}\vec{e}_{p} \\
    \f{\world}\vec{e}_{p,i} \\
    \f{\world}\vec{e}_{v} \\
    \f{\world}\vec{e}_{a} \\
    \f{\body}\vec{e}_{R} \\
    \f{\body}\vec{e}_{R,i} \\
    \f{\body}\vec{e}_{\omega} \\
    \f{\body}\vec{e}_{\psi}
    \end{bmatrix}
    =
    \begin{bmatrix}
    \f{\world}\vec{p}-\f{\world}\vec{p}_{\des} \\
    \int\f{\world}\vec{e}_{p}\diff t \\
    \f{\world}\vec{v}-\f{\world}\vec{v}_{\des} \\
    \f{\world}\vec{a}-\f{\world}\vec{a}_{\des} \\
     \frac{1}{2}\left( \rot{\world}{\body_{\des}}^T\rot{\world}{\body}-\rot{\world}{\body}^T\rot{\world}{\body_{\des}}\right)^\vee \\
    \int\f{\body}\vec{e}_{R}\diff t\\
    \f{\body}\vec{\omega}_{\world\body}-\rot{\body}{\world}\f{\world}\vec{\omega}_{\world\body_{\des},\des} \\
    \f{\body}\vec{e}_{\psi}
    \end{bmatrix}
\end{equation}
with $\vec{e}\in\mathbb{R}^{24}$.
Note that the angular velocity reference $\f{\world}\vec{\omega}_{\world\body_{\des},\des}$ denotes the reference angular velocity of the reference body frame with respect to the world frame.
The geometric attitude error

The error dynamics are derived as:
\begin{subequations}
\begin{align}
    \f{\world}\dot{\vec{e}}_{p} &= \f{\world}\vec{e}_{v}\\
    \f{\world}\dot{\vec{e}}_{p,i} &= \f{\world}\vec{e}_{p}\\
    \f{\world}\dot{\vec{e}}_{v} &= \f{\world}\vec{e}_{a}\\
     \f{\world}\dot{\vec{e}}_{a} &=
    \begin{multlined}[t]
    \frac{1}{m}\bm{R}_{\world\body}\left(\f{\body}(\dot{\vec{f}})-[\f{\body}\vec{\omega}_{\world\body}]_\times \bm{R}_{\body\world}m\f{\world}\vec{g} \right)\\ - \f{\world}\vec{j}_{\des}
    \end{multlined}\label{eq:lin_jerk_error}\\
    \f{\body}\dot{\vec{e}}_{R} &= 
    \underbrace{
    \frac{1}{2}\left(\tr\left( \bm{R}_{\world\body}^T \bm{R}_{\body\world_{\des}} \right)\eye{3} - \bm{R}_{\world\body}^T\bm{R}_{\world\body_{\des}}\right)}_{\bm{A}_{R\Omega}(\bm{R}_{\world\body},\bm{R}_{\world\body_d})}
    \f{\body}\vec{e}_\omega\\
    \f{\body}\dot{\vec{e}}_{R,i} &= \f{\body}\vec{e}_{R}\\
    \f{\body}\dot{\vec{e}}_{\omega} &= \f{\body}\vec{e}_{\psi}\\
    \f{\body}\dot{\vec{e}}_{\psi} &{}\label{eq:ang_jerk_error}
\end{align}
\end{subequations}
Refer to \cref{sec:appendix_jerk}, \cref{eq:error_dynamics_angacc} for a full formulation of the angular acceleration error dynamics.

\subsubsection{Feedback linearization}
We use feedback linearization to simplify the error dynamics, specifically \cref{eq:lin_jerk_error} and \cref{eq:ang_jerk_error}. We introduce the virtual input vector $\bar{\vec{u}}\in\mathbb{R}^6$  and define the total force derivative as follows:
\begin{equation}\label{eq:lin_fbl2}
\begin{split}
\f{\body}(\dot{\vec{f}}) &= [\f{\body}\vec{\omega}_{\world\body}]_\times \bm{R}_{\body\world}m\f{\world}\vec{g} 
    + m \bm{R}_{\world\body} \f{\world}\vec{j}_\des\\
    &+ m \bm{R}_{\world\body} \begin{bmatrix} \bar{u}_1\\ \bar{u}_2\\ \bar{u}_3\end{bmatrix}
\end{split}
\end{equation}
We follow the same principle for the angular dynamics, the full derivation is shown in \cref{sec:appendix_linearization}, \cref{eq:ang_fbl2}.

Using the definitions from \cref{eq:lin_fbl2} and \cref{eq:ang_fbl2} together with \cref{eq:lin_jerk_error} and \cref{eq:ang_jerk_error} we obtain the simplified error dynamics for linear and angular acceleration:
\begin{equation}\label{eq:feedback_linearized_system}
    \f{\world}\dot{\vec{e}}_{a} = \begin{bmatrix} \bar{u}_1\\ \bar{u}_2\\ \bar{u}_3\end{bmatrix} \qquad
    \f{\body}\dot{\vec{e}}_{\psi} = \begin{bmatrix} \bar{u}_4\\ \bar{u}_5\\ \bar{u}_6\end{bmatrix}
\end{equation}

\subsubsection{Linearization}
Having performed the feedback linearization, we obtain following nonlinear error dynamics:
\begin{equation}
\begin{aligned}
    \label{eq:nonlin}
    \dot{\vec{e}} &= f(\vec{e},\bar{\vec{u}})\\
    \frac{d}{dt}
    \begin{bmatrix}
    \f{\world}\vec{e}_{p}\\ \f{\world}\vec{e}_{p,i}\\ \f{\world}\vec{e}_{v}\\ \f{\world}\vec{e}_{a} \\
    \f{\body}\vec{e}_{R}\\ \f{\body}\vec{e}_{R,i}\\ \f{\body}\vec{e}_{\omega} \\ \f{\body}\vec{e}_{\psi}
    \end{bmatrix}
    &=
    \begin{bmatrix}
    \f{\world}\vec{e}_{v}\\ 
    \f{\world}\vec{e}_{p}\\ 
    \f{\world}\vec{e}_{a}\\
    \begin{bmatrix} \bar u_1 & \bar u_2 & \bar u_3 \end{bmatrix}^T\\
    \bm{A}_{R\Omega}(\bm{R}_{\world\body},\bm{R}_{\world\body_{\des}})\f{\body}\vec{e}_{\omega}\\
    \f{\body}\vec{e}_{R}\\ 
    \f{\body}\vec{e}_{\psi}\\ 
    \begin{bmatrix} \bar u_4 & \bar u_5 & \bar u_6 \end{bmatrix}^T
    \end{bmatrix}
\end{aligned}
\end{equation}

As the attitude error dynamics are nonlinear, we linearize them at each time step around zero attitude error, i.e. $\f{\body}\vec{e}_{R}=\vec{0}$, to obtain a linear representation of the form
\begin{equation}
\label{eq:dynamic_matrix}
    \dot{\vec{e}} = \bm{A}\vec{e} + \bm{B}\bar{\vec{u}}\quad \bm{A}\in\mathbb{R}^{24\times24},\bm{B}\in\mathbb{R}^{24\times6}
\end{equation}

Where $\bm{A}$ is the state transition matrix and $\bm{B}$ is the input matrix.

\subsubsection{Optimal gain computation}
The optimal control inputs can then be computed using the gain $\bm{K}_{LQRI}$:
\begin{equation}\label{eq:lqr_control}
    \bar{\vec{u}}=-\bm{K}_{LQRI}\vec{e}\qquad \bar{\vec{u}}\in\mathbb{R}^6,\ \bm{K}_{LQRI}\in\mathbb{R}^{6\times24}
\end{equation}
To find the optimal \ac{LQRI} gain matrix, we solve the continuous time algebraic Riccati equation (CARE):
\begin{equation}
    \label{eq:care}
    0 = \bm{A}^T\bm{P} + \bm{PA} - \bm{PBR}^{-1}\bm{B}^T\bm{P} + \bm{Q}
\end{equation}
and then use it's solution $\bm{P}\succ0,\ \bm{P}\in\mathbb{R}^{24\times24}$ according to:
\begin{equation}
    \label{eq:lqri1_gain}
    \bm{K}_{LQRI} = \bm{R}^{-1}\bm{B}^T\bm{P}
\end{equation}

A stability proof can be found in \cref{sec:stability_proof}.

\subsection{PID controller}
In this section we present a \ac{PID} controller that yields rigid body jerk commands. The jerk commands are computed by numerically differentiating acceleration commands that are computed from a \ac{PID} controller as in \cite{bodie2018towards,kamel2018voliro}.

Using the current as well as the reference state of the system, reference accelerations are obtained as follows:
\begin{subequations}
\begin{align}
    \f{\world}\vec{a} &= \bm{K}_{p}\f{\world}\vec{e}_{p} + \bm{K}_{v}\f{\world}\vec{e}_{v} + \bm{K}_{p,i}\f{\world}\vec{e}_{p,i}\\
    \f{\body}\vec{\psi} &= \bm{K}_{R}\f{\body}\vec{e}_{R} + \bm{K}_{\omega}\f{\body}\vec{e}_{\omega} + \bm{K}_{R,i}\f{\body}\vec{e}_{R,i}
\end{align}
\end{subequations}
The error terms are defined as in \cref{eq:error_terms} and the matrices $\bm{K}_{(\cdot)}$ are diagonal tuning matrices.
The reference accelerations are then numerically differentiated and rotated to obtain the body jerk commands:
\begin{equation}
    \vec u = \begin{bmatrix}\f{B}\vec{j} \\ \f{B}\vec{\zeta}\end{bmatrix}
    =\begin{bmatrix}
        \rot{\body}{\world}\f{\world}\dot{\vec{a}} \\
        \f{\body}\dot{\vec{\psi}}
    \end{bmatrix}
\end{equation}

\subsection{Differential actuator allocation}
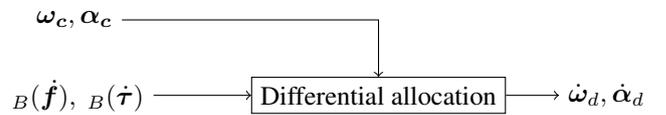
\begin{figure}[h]
 \centering
\tikzstyle{block} = [rectangle, text centered, draw=black]
\tikzstyle{arrow} = [thick,->,>=stealth]
\begin{tikzpicture}
\node (current_inputs) {$\vec{\omega_c},\vec{\alpha_c}$};
\node (df) at (0,-1) {$\f{\body}(\dot{\vec{f}}) ,\f{\body}(\dot{\vec{\tau}})$};
\node(diff_alloc)[block] at (4,-1) {Differential allocation};
\node(result) at (7,-1) {$\dot{\vec{\omega}}_d,\dot{\vec{\alpha}}_d$};
\draw[->](current_inputs)-|(diff_alloc);
\draw[->](df)--(diff_alloc);
\draw[->](diff_alloc)--(result);
\end{tikzpicture}
 \caption{Differential allocation.}
 \label{fig:dF_block}
\end{figure}

The LQRI and PID controllers presented above return jerk commands that need to be executed by the platform.
Extending the work from \cite{bodie2018towards} and based on work by \cite{ryll2015novel}, we present a weighted differential allocation of actuator commands, and include task prioritization in the null space.

Taking the time derivative of the standard allocation, we obtain the relation between the derivative of the body wrench, actuator controls and their derivatives:
\begin{equation}
\begin{aligned}
    \frac{d}{dt}&\left(
    \begin{bmatrix}\f{\body}\vec{f}\\
    \f{\body}\vec{\tau}\\
    \end{bmatrix}\right)
    = \frac{d}{dt}\left( \bm{A}\tilde{\vec{\Omega}} \right)\\
    &=
    \frac{d}{dt}\left(
    \bm{A}
    \begin{bmatrix}
        \omega_{1}^2\sin(\alpha_{c,1})\\
        \omega_{1}^2\cos(\alpha_{c,1})\\
        \vdots\\
        \omega_{12}^2\sin(\alpha_{c,6})\\
        \omega_{12}^2\cos(\alpha_{c,6})\\
    \end{bmatrix}\right)\\
&=\underbrace{\bm{A}\cdot\Delta \bm{A}(\vec{\omega}_c,\vec{\alpha}_c)}_{\tilde{\bm{A}}}\cdot
    \underbrace{\begin{bmatrix}
        \dot{\omega}_{d,1} \\
         \vdots\\
        \dot{\omega}_{d,12} \\
         \dot{\alpha}_{d,1}\\
         \vdots\\
         \dot{\alpha}_{d,6}\\
     \end{bmatrix}}_{\tilde{\vec{u}}\in\mathbb{R}^{18}}
     \end{aligned}\label{eq:diff_allocation}
\end{equation}
The vectors $\vec{\omega}_c, \vec{\alpha}_c$ contain the current actuator commands, and the vector $\tilde{\vec{u}}=\begin{bmatrix}\dot{\vec{\omega}}^T_d &\dot{\vec{\alpha}}^T_d  \end{bmatrix}^T$ represents the desired derivatives of the actuator commands, i.e. rotor accelerations and tilt angle velocities.

Note that \cref{eq:diff_allocation} requires the derivative of the body \emph{wrench} as input, while the controller returns a desired body \emph{jerk}. Just like on the acceleration level, these two quantities are related by the body inertia properties:
\begin{equation}
    \begin{bmatrix}
        \f{\body}(\dot{\vec{f}})\\
        \f{\body}(\dot{\vec{\tau}})\\
    \end{bmatrix}=
    \begin{bmatrix}
    m\eye{3} & \bm{0} \\
    \bm{0} & \inertia \\
    \end{bmatrix}
    \begin{bmatrix}
        \f{\body}\vec{j}\\
        \f{\body}\vec{\zeta}\\
    \end{bmatrix}
\end{equation}
Since the differential allocation matrix $\tilde{\bm{A}}$ is in $\mathbb{R}^{6\times18}$, we have a high dimensional nullspace that can be exploited when computing $\tilde{\vec{u}}$ using its pseudoinverse. In order to gain control over this freedom, we formulate following optimization problem:
\begin{equation}
\begin{aligned}
    \min_{\tilde{\vec{u}}} &\ \norm{\bm{W}\left(\tilde{\vec{u}}-\tilde{\vec{u}}^*\right)}_2\\
    \text{s.t.:}&\ 
    \begin{bmatrix}
        \f{\body}(\dot{\vec{f}})\\
        \f{\body}(\dot{\vec{\tau}})\\
    \end{bmatrix}
    =
    \tilde{\bm{A}}\tilde{\vec{u}},
\end{aligned}
\label{eq:allocation_optimization}
\end{equation}
where $\bm{W}\in\mathbb{R}^{18\times18}$ is a weighting matrix and $\tilde{\vec{u}}^*$ are optimal rotor acceleration and tilt velocity values. The analytical solution to the problem is given as:
\begin{equation}\label{eq:diff_input}
\tilde{\vec{u}}=
     \tilde{\vec{u}}^* + \bm{W}\tilde{\bm{A}}^T\left( \tilde{\bm{A}}\bm{W}\tilde{\bm{A}}^T \right)^{-1}
     \left(
    \begin{bmatrix}
        \f{\body}(\dot{\vec{f}})\\
        \f{\body}(\dot{\vec{\tau}})\\
    \end{bmatrix}
    - \tilde{\bm{A}} \tilde{\vec{u}}^*\right)
\end{equation}
Using the optimization from \cref{eq:allocation_optimization}, we are able to choose optimal differential controls $\tilde{\vec{u}}^*$ freely.
Here we present our method of finding a $\tilde{\vec{u}}^*$ that achieves an efficient control configuration and that allows arm unwinding mid-flight.

Based on the current control inputs $\vec{\alpha}_{c}$ and $\vec{\omega}_{c}$, we compute the resulting body wrench. Using the pseudo inverse of the static allocation matrix (\cref{eq:static_allocation}), we can compute the optimal tilt angles $\vec{\alpha}^*$ and rotor speeds $\vec{\omega}^*$. The optimal differential controls $\tilde{\vec{u}}^*$ are then computed from the difference and are assigned unwinding velocities $v_{\dot{\omega}}$ and $v_{\dot{\alpha}}$:
\begin{subequations}
\begin{align}
                \dot{\vec{\omega}}^* &= \sign\left(\vec{\omega}^* - \vec{\omega}_{c}\right)v_{\dot{\omega}}\label{eq:unwind_o}\\
                \dot{\vec{\alpha}}^* &= \sign\left(\vec{\alpha}^* - \vec{\alpha}_{c}\right)v_{\dot{\alpha}}\label{eq:unwind_a}
\end{align}
\end{subequations}
\begin{figure}[h]
 \centering
\includegraphics[width=\columnwidth]{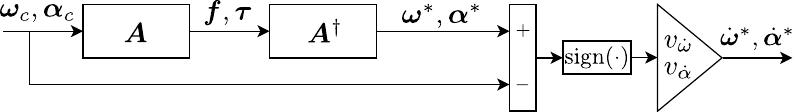}
 \caption{Optimal control allocation.}
 \label{fig:control_optimal_allocation}
\end{figure}

After having computed the optimal control inputs we use \cref{eq:diff_input} to find the desired differential control inputs $\tilde{\vec{u}}$. In a next step they are saturated according to predefined limits on maximum rotor and tilt angle velocities and subsequently integrated over time (see \cref{fig:complete_control_diagram}). The resulting commands for tilt angles $\vec{\alpha}$ and for rotor speeds $\vec{\omega}$ are then sent to the actuators.


Kinematic singularities are handled inherently by the inclusion of $\dot{\bm{\alpha}}$ in the differential actuator allocation. Rank reduction singularities are still present, and we evaluate them by computing the condition number of the instantaneous allocation matrix, $\kappa(\bm{A}_{\alpha})$, defined as

\begin{equation}
    \kappa(\bm{A}_{\alpha}) = \frac{\sigma_{max}(\bm{A}_{\alpha})}{\sigma_{min}(\bm{A}_{\alpha})}
\end{equation}

where $\sigma_{max}$ and $\sigma_{min}$ are the maximum and minimum singular values of $\bm{A}_{\alpha}$.

\begin{figure*}[t]
 \centering
\includegraphics[width=\textwidth]{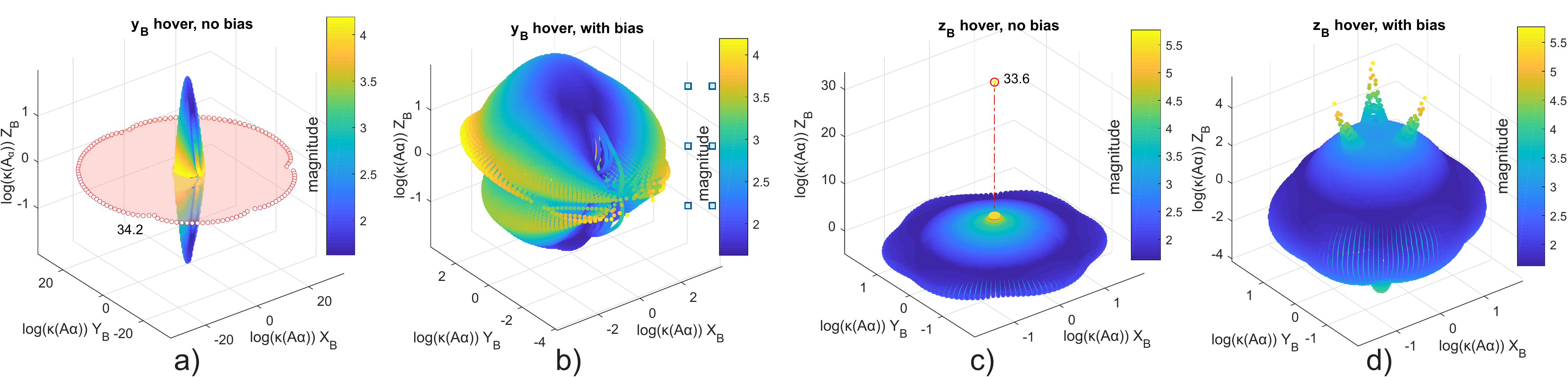}
 \caption{The log of the condition number $\kappa(\bm{A}_{\alpha})$ is plotted in $\frame{\body}$ for the cases of $y_{\body}$ (a,b) and $z_{\body}$ (c,d) hover with an envelope of additional gravitational force in all directions. Envelopes with a tilt bias compensation on $\bm{\alpha}$ (b,d) significantly reduce the maximum condition numbers. Condition number magnitudes are indicated on the color scale, with $\kappa_{max}(\bm{A}_{\alpha})$ indicated in (a,c)}
 \label{fig:condition_num}
\end{figure*}

\Cref{fig:condition_num} plots the log of $\kappa(\bm{A}_{\alpha})$ over a force envelope of magnitude $mg$ in additional to a static hover force in $y_{\body}$ and $z_{\body}$ directions. An allocation of the current wrench according to \cref{eq:static_allocation} produces the envelopes shown in a) and c), with high condition numbers in the $z_{\body}$-plane, and along the $z_{\body}$-axis respectively. High condition numbers occur where $\sigma_{min}(\bm{A}_{\alpha})$ is significantly reduced, indicating loss of rank and therefore loss of instantaneous controllability in at least one \ac{DOF}.

We can address the above-mentioned rank reduction in the formulation of $\vec{\alpha}^*$, incorporating a bias term $\bm{\alpha}_{bias}$ derived according to the alignment of $\bm{f}_{d}$ in the body frame. The detailed derivation of $\bm{\alpha}_{bias}$ is presented by \cite{bodie2018towards}. The addition of this bias term to $\vec{\alpha}^*$ results in the $\text{log}(\kappa(\bm{A}_{\alpha}))$ envelopes in \cref{fig:condition_num}(b,d). In these plots, values of $\kappa_{max}(\bm{A}_{\alpha})$ are significantly reduced, from $\text{exp}(34.2)$ to $\text{exp}(4.19)$ in $y_{\body}$ hover, and $\text{exp}(33.6)$ to $\text{exp}(5.77)$ in $z_{\body}$ hover.

\section{Experimental Setup}
\label{sec:setup}
\subsection{System hardware}
\label{sec:platform}
\begin{figure}[htp]
 \centering
\includegraphics[width=\columnwidth]{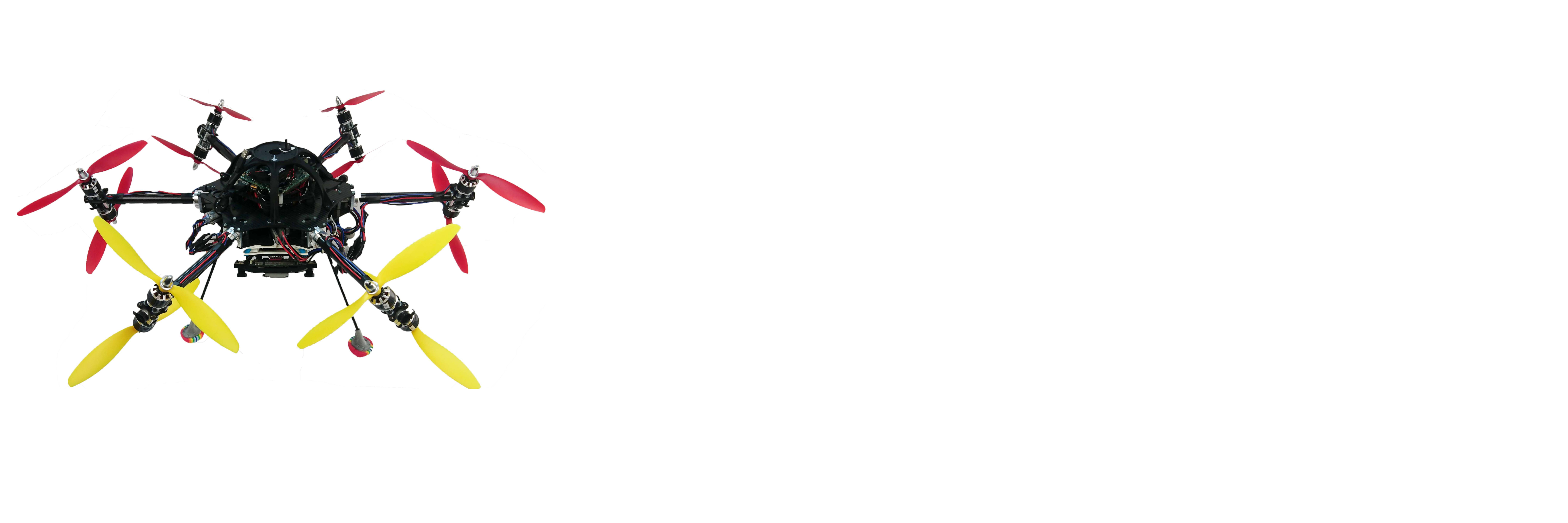}
 \caption{Image of the prototype system.}
 \label{fig:prototype}
\end{figure}

The experimental platform modeled after the morphology presented in \cref{sec:design} is a 12 rotor \ac{MAV} with 6 tiltable arms, shown in \cref{fig:prototype}. Primary system properties and components are reported in \cref{tab:prototype}, and described below.
The system layout consists of equally spaced arms along the $z_b$-plane, with two rotors per arm to balance the rotational inertia.
Two KDE 2315XF-885 motors per arm with $9$in (\SI{228.6}{\milli\meter}) propellers provide sufficient thrust (\SI{11}{\newton} per motor), to enable dynamic flight in the least efficient configurations.
Each arm is independently tilted by a Dynamixel XL430 servo actuator, located in the base to reduce system inertia.
Upper and lower propellers counter-rotate such that drag torques from the propellers are approximately canceled, and the gyroscopic moment on the tilting mechanism is minimized, reducing the effort of the tilt motor.
The controller and state estimator operate on an onboard Intel NUC i7, which sends direct actuator commands to a Pixhawk flight controller. All rotors and tilting servos, as well as a high resolution \ac{IMU}, are connected to the flight controller.
The system is powered by two 3800 mAh 6S LiPo batteries.
The total system mass is \SI{4.27}{\kilogram}, and it can generate over \SI{130}{\newton} of force at maximum thrust in the horizontal hover configuration.

\begin{table}[]
\caption{Main system components and parameters}
\label{tab:prototype}
\begin{threeparttable}

\centering
\def\arraystretch{1.2}

\begin{tabularx}{\columnwidth}{l c c}
\hline
component & part number & qty \\
\hline
onboard computer & Intel NUC i7 & 1 \\
flight controller & Pixhawk mRo & 1\\
rotor & KDE 2315XF-885 & 12 \\
propeller & Gemfan 9x4.7 & 12 \\
ESC & T-motor F45A 3-6S & 12 \\
tilt motor & Dynamixel XL430-W250 & 6 \\
IMU & ADIS16448 & 1 \\
battery & 3800 mAh 6S LiPo & 2 \\
\hline
\end{tabularx}

\begin{tabularx}{\columnwidth}{l c c}
\hline
parameter & value & units\\
\hline
rotor groups & 6 & \\
arm length, $l$ & 0.3 & [\si{\meter}] \\ 
total mass & 4.27 & [\si{\kilogram}] \\
inertia$^{*}$ & $\{0.086, 0.088, 0.16\}$ & [\si{\kilogram}\si{\meter}$^2$] \\
diameter & 0.83 & [\si{\meter}] \\
$f_{\rotor{i},max}$ & 11 & [\si{\newton}] \\
$c_f$ & $7.1\text{e}{-6}$ & [\si{\newton}\si{\second}$^2$/\si{\radian}$^2$] \\
$\omega_{max}$ & 1250 & [\si{\radian}/\si{\second}] \\
\hline
\end{tabularx}
\begin{tablenotes}
\item[] $^{*}$Primary components of inertia are obtained from \ac{CAD} model.
\end{tablenotes}
\end{threeparttable}
\end{table}

\subsection{Test environment}
\begin{figure}[h]
 \centering
\includegraphics[width=\columnwidth]{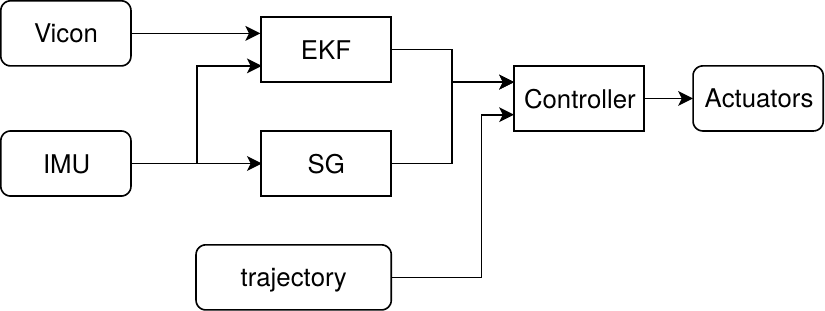}
 \caption{System setup: Vicon and \ac{IMU} measurements are fused by an \ac{EKF}, and combined with the Savitzky-Golay (SG) filtered angular acceleration estimate for complete state estimation. The controller processes the state estimate and reference trajectory, and sends actuator commands.}
 \label{fig:system_block_diagram}
\end{figure}

All experiments are performed in a laboratory environment, where an external Vicon motion capture system wirelessly sends high-accuracy odometry information to the onboard computer. 
The test area provides a space of approximately 4x4 meters in size and 3 meters in height.
Motion capture pose estimates are taken at \SI{20}{\hertz} and fused by an \ac{EKF} with the onboard \ac{IMU} at a rate of \SI{102}{\hertz}.
Reference trajectories are sent to the controller from a ground control station, or from the onboard computer. The trajectories are polynomials that are generated offline based on discrete waypoints. They contain smooth reference commands for position, velocity, acceleration, as well as attitude, angular velocity, and angular acceleration, and are sampled at \SI{100}{\hertz}.
A block diagram of the experimental system setup is shown in \cref{fig:system_block_diagram}).

\subsection{Acceleration estimation}
The suggested jerk level \ac{LQRI} requires linear and angular acceleration estimates to find an optimal gain matrix. We use a first-order \ac{SG} filter to smooth raw IMU acceleration and angular velocity data. The smoothed angular velocity data is differentiated numerically to obtain angular acceleration estimates.

\subsection{Controller parameters}
Controller parameters and gains used in experimental tests are listed in \cref{tab:experimental_params}.

\begin{table}[]
\caption{Controller parameters for experimental tests}
\label{tab:experimental_params}
\centering
\def\arraystretch{1.2}
\begin{tabularx}{\columnwidth}{l c | l c}
\hline
parameter & value & parameter & value\\
\hline
\textbf{LQRI} & & \textbf{PID} & \\
\hline
$k_{p}$ & 200 & $k_{p}$ & 5 \\
$k_{p,i}$ & 50 & $k_{p,i}$ & 0.3 \\
$k_{v}$ & 100 & $k_{v}$ & 1.0 \\
$k_{a}$ & 0 & $k_{R}$ & 3.5 \\
$k_{R}$ & 100 & $k_{R,i}$ & 0.3 \\
$k_{R,i}$ & 100 & $k_{\omega}$ & 0.8 \\
\cmidrule{3-4}
$k_{\omega}$ & 200 & \textbf{Allocation} & \\
\cmidrule{3-4}
$k_{\psi}$ & 0 & $k_{\alpha}$ & 1000 \\
$r_{\dot{f}}$ & $(1.0,1.0,0.2)$ & $v_{\alpha}$ & 1 \\
$r_{\dot{\tau}}$ & $(1.0,1.0,1.0)$ & $v_{\omega}$ & 250 \\
\hline

\end{tabularx}
\end{table}

\section{Experimental Results}
\label{sec:experiments}
\subsection{Trajectories and data collection}
In order to evaluate and to compare tracking performance of the controllers, several experimental trajectories are designed and tested. Videos of the experimental trials can be found in our supporting multimedia content.

\begin{figure}[h]
 \centering
\includegraphics[width=\columnwidth]{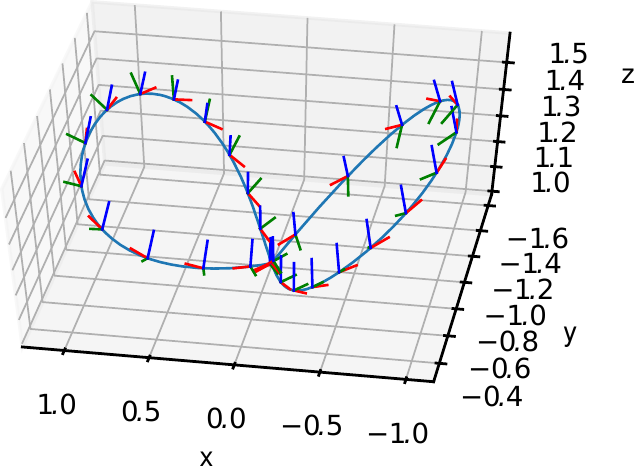}
 \caption{Reference figure eight trajectory used for performance evaluation.}
 \label{fig:reference_trajectory}
\end{figure}

We first define a reference trajectory that covers large parts of the 6-dimensional pose space. The trajectory can be described as a figure eight with varying height and attitudes. \Cref{fig:reference_trajectory} illustrates the trajectory with some reference positions and attitudes.
We use three different adaptions of the trajectory to evaluate controller performance, specifically:
\begin{itemize}
    \item Trajectory (a), low angle: The maximum tilt angle does not exceed 30 degrees, the duration of the trajectory is 29.4 seconds.
    \item Trajectory (b), high angle: The maximum reference tilt angle does not exceed 80 degrees and the duration is 29.4 seconds.
    \item Trajectory (c), fast tracking: Using a maximum tilt angle of 30 degrees, the duration of the trajectory is 10.7 seconds.
\end{itemize}


Another set of trajectories target the singular configurations. The first commands a lateral translation and rotation while in both a kinematic and rank reduced singular state, namely at 90 degrees roll, and a second demonstrates transition through the kinematic singularity.

\begin{itemize}
    \item Trajectory (d), singular translation: The trajectory transitions to a \SI{90}{\degree} roll, translates, reverses direction, and rotates about the $y_{\body}$-axis. The duration of the trajectory is \SI{36.1}{\second}.
    \item Trajectory (e), cartwheel: The trajectory transitions to a \SI{90}{\degree} roll, then makes two complete rotations about the $z_{\body}$-axis while translating in a circular trajectory. The duration of the trajectory is \SI{35.5}{\second}.
\end{itemize}

Two additional trajectories are designed to evaluate the secondary unwinding task with a complete rotation about a specified axis:

\begin{itemize}
    \item Trajectory (f), roll flip: complete rotation about the $x_{\body}$ axis. The duration of the trajectory is \SI{16}{\second}.
     \item Trajectory (g), pitch flip: complete rotation about the $y_{\body}$ axis. The duration of the trajectory is \SI{8}{\second}.
\end{itemize}

Position and attitude errors throughout the section are taken from data collected over the complete duration of a single iteration of each trajectory, for both LQRI and PID controllers. Since we are interested in exploiting the omnidirectional capabilities of the system and thus track all 6 axes of a trajectory, errors are given separately in all axes.

Position errors are given in the world frame in meters, attitude errors are given with respect to the reference attitude in radians.

All errors are evaluated in boxplots, showing the median, the upper and lower quartile, as well as the 1.5 \ac{IQR}.



\begin{figure}[h]
 \centering
\includegraphics[width=\columnwidth]{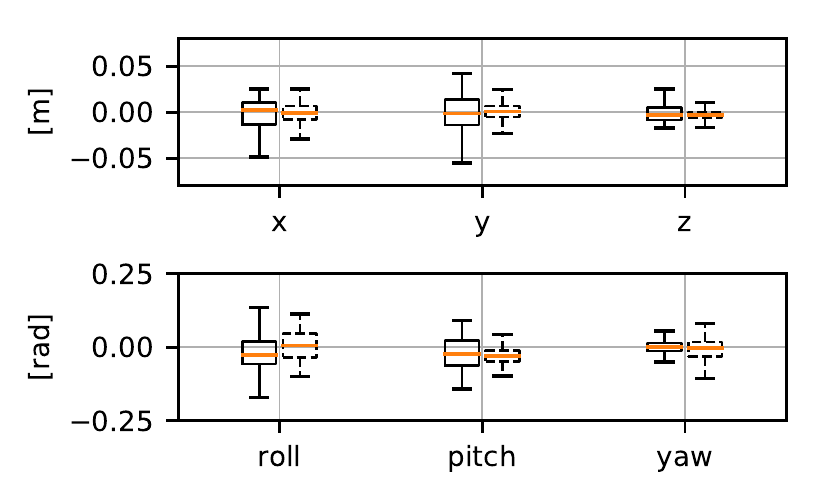}
 \caption{Position and attitude errors of LQRI (solid) and PID (dashed) for trajectory (a), low angle figure eight.}
 \label{fig:lqri_pid_pos_att_errors_standard}
\end{figure}

\begin{figure}[h]
 \centering
\includegraphics[width=\columnwidth]{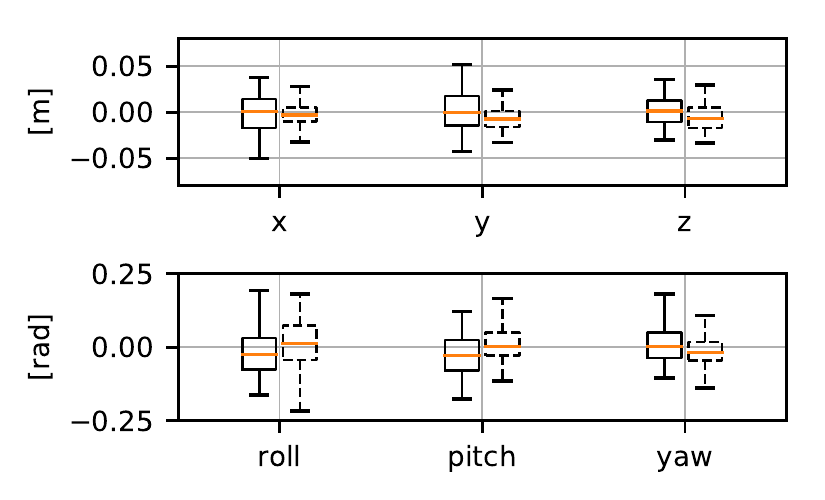}
 \caption{Position and attitude errors of LQRI (solid) and PID (dashed) for trajectory (b), high angle figure eight.}
 \label{fig:lqri_pid_pos_att_errors_steep}
\end{figure}

\begin{figure}[h]
 \centering
\includegraphics[width=\columnwidth]{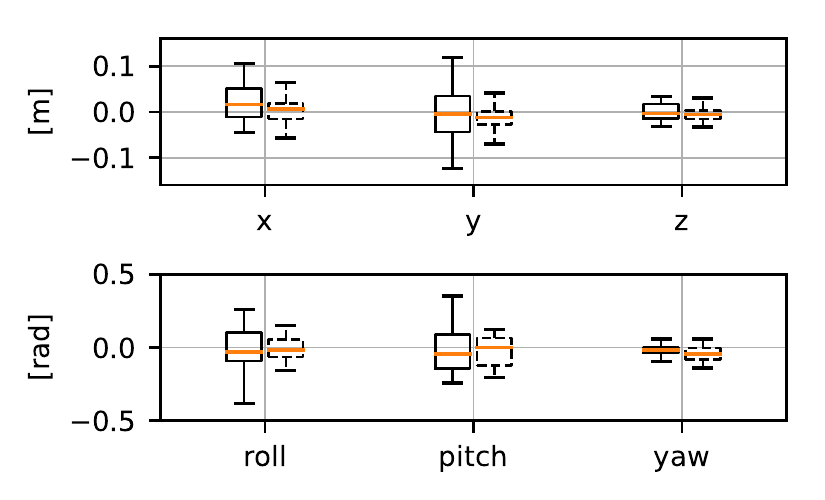}
 \caption{Position and attitude errors of LQRI (solid) and PID (dashed) for trajectory (c), fast figure eight.}
 \label{fig:lqri_pid_pos_att_errors_fast}
\end{figure}

\begin{figure}[h]
 \centering
\includegraphics[width=\columnwidth]{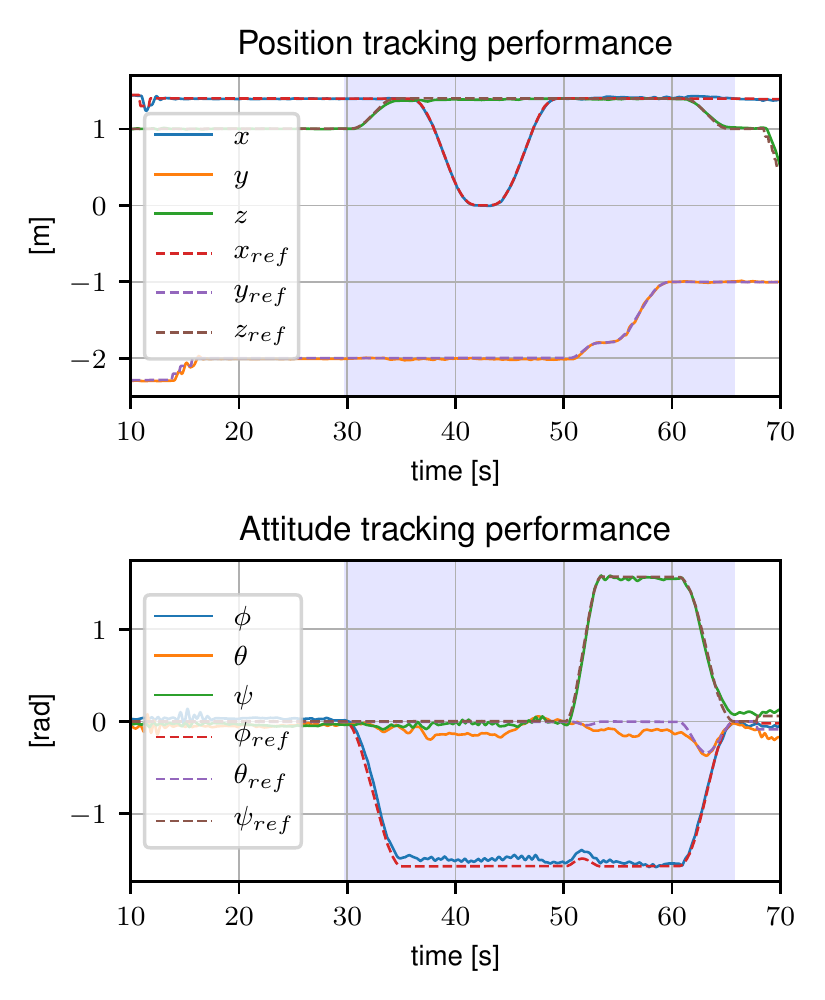}
 \caption{Position and attitude errors of LQRI (solid) and PID (dashed) for trajectory (d), singular translation.}
 \label{fig:singularity_reference_tracking}
\end{figure}

\begin{figure}[h]
 \centering
\includegraphics[width=\columnwidth]{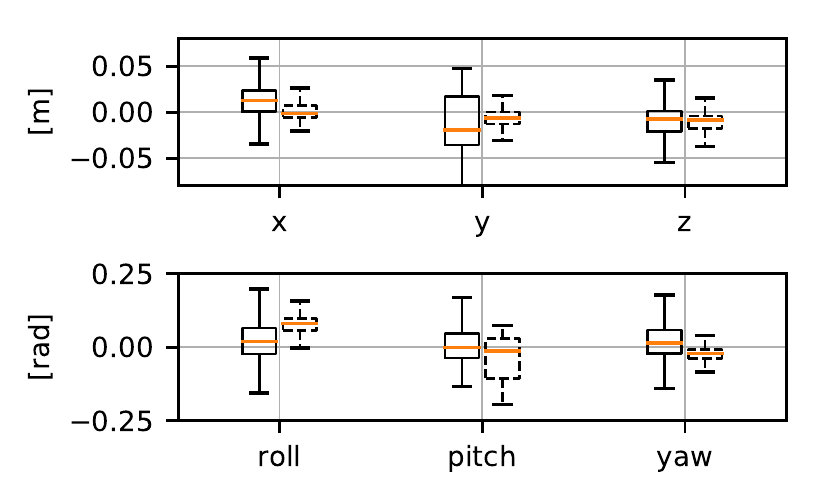}
 \caption{LQRI vs PID tracking errors during trajectory (d), singular translation.}
 \label{fig:lqri_pid_singularity_tracking_error}
\end{figure}

\begin{figure}[h]
 \centering
\includegraphics[width=\columnwidth]{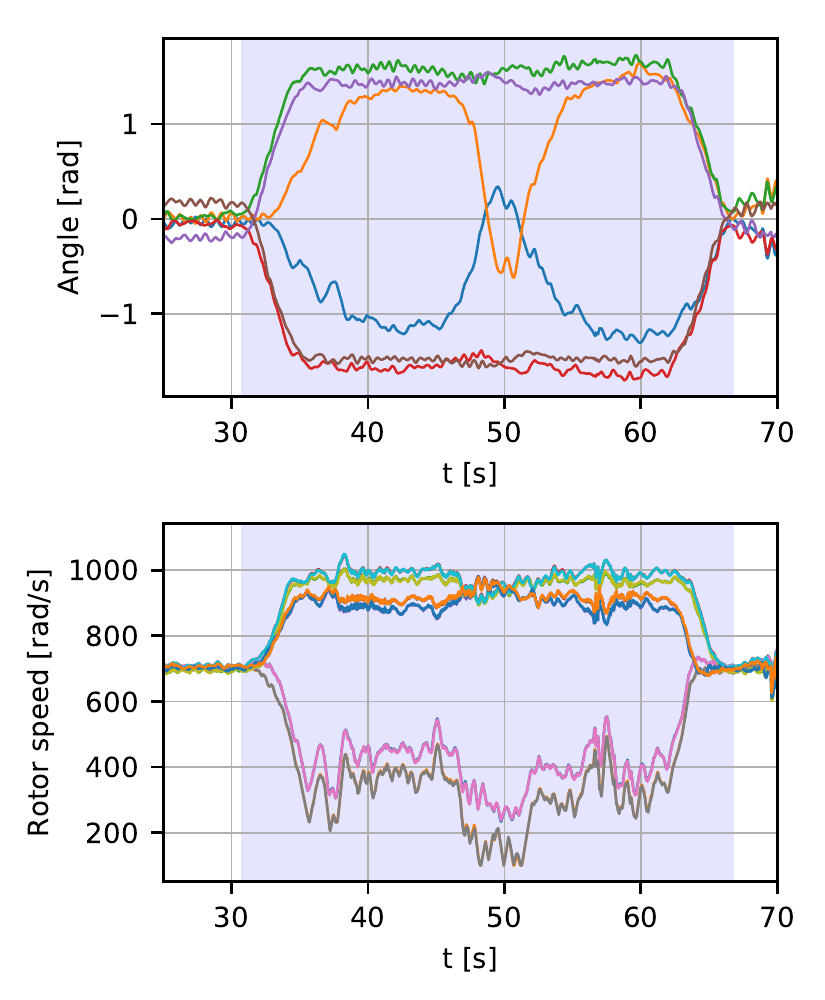}
 \caption{PID direct controls during trajectory (d), singular translation.}
 \label{fig:singularity_tracking_direct_controls}
\end{figure}

\begin{figure}[h]
 \centering
\includegraphics[width=\columnwidth]{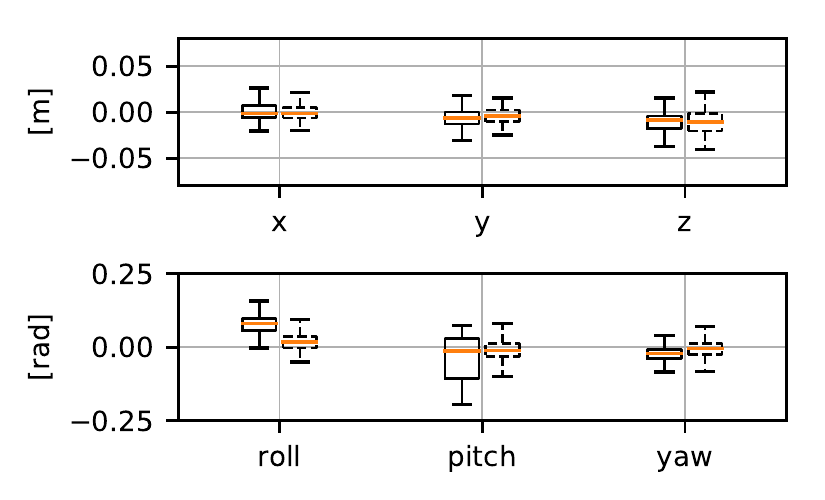}
 \caption{Tracking accuracy of PID with and without singularity handling, of trajectory (d), singular translation.}
 \label{fig:singularity_tracking_offsets}
\end{figure}

\begin{figure}[h]
 \centering
\includegraphics[width=\columnwidth]{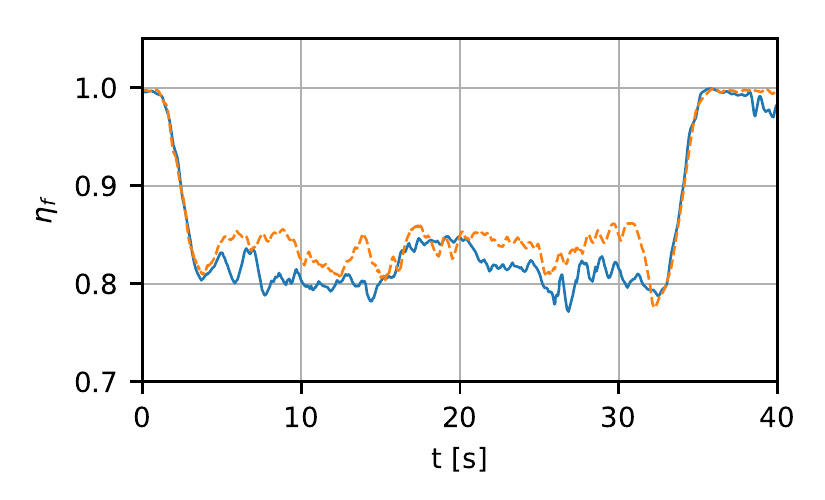}
 \caption{Force efficiency index of PID, with (solid) and without (dashed) singularity handling, of trajectory (d), singular translation.}
 \label{fig:singularity_wfi}
\end{figure}


\begin{figure}[h]
 \centering
\includegraphics[width=\columnwidth]{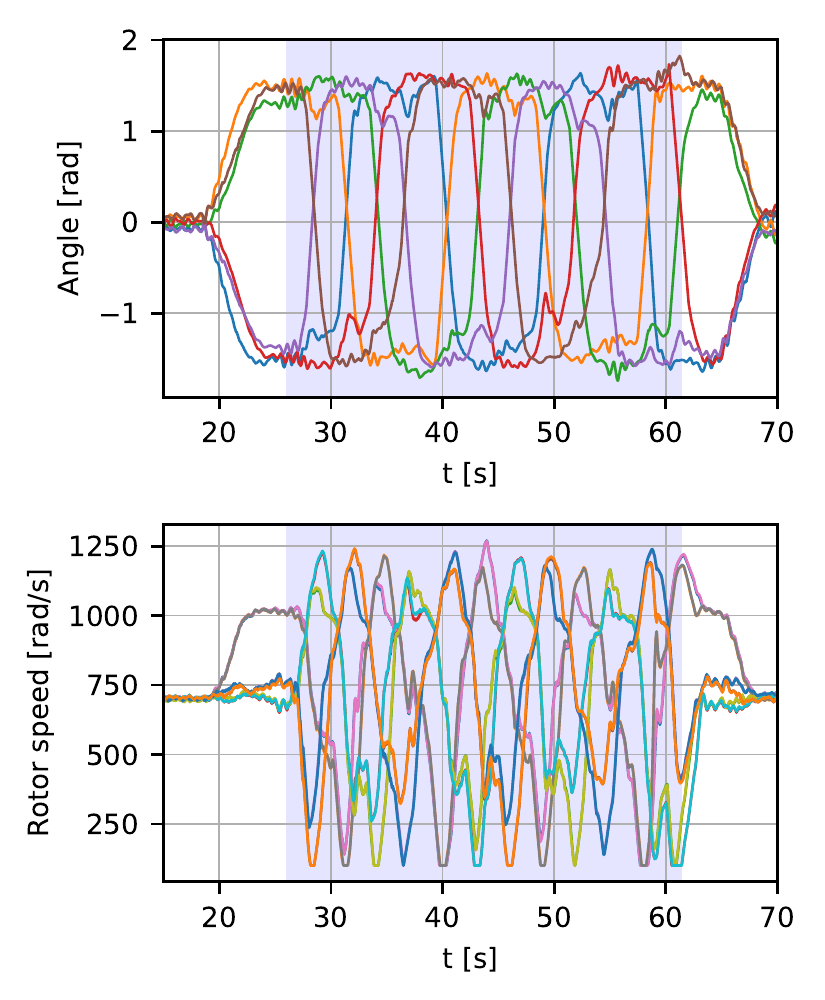}
 \caption{Trajectory (e) actuator commands during a cartwheel.}
 \label{fig:cartwheel_direct_control_inputs}
\end{figure}


\begin{figure}[h]
 \centering
\includegraphics[width=\columnwidth]{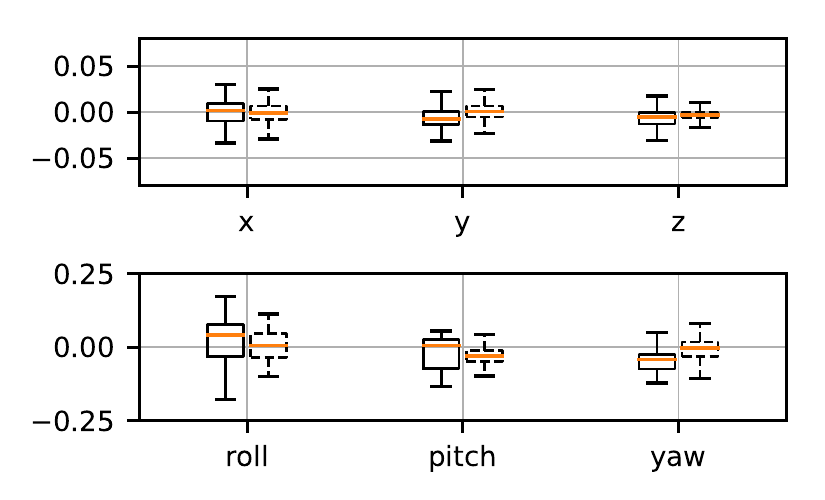}
 \caption{Trajectory (a) tracking errors with and without unwinding arms (solid and dashed lines, respectively).}
 \label{fig:unwinding_standard_trajectory}
\end{figure}

\begin{figure}[h]
 \centering
\includegraphics[width=\columnwidth]{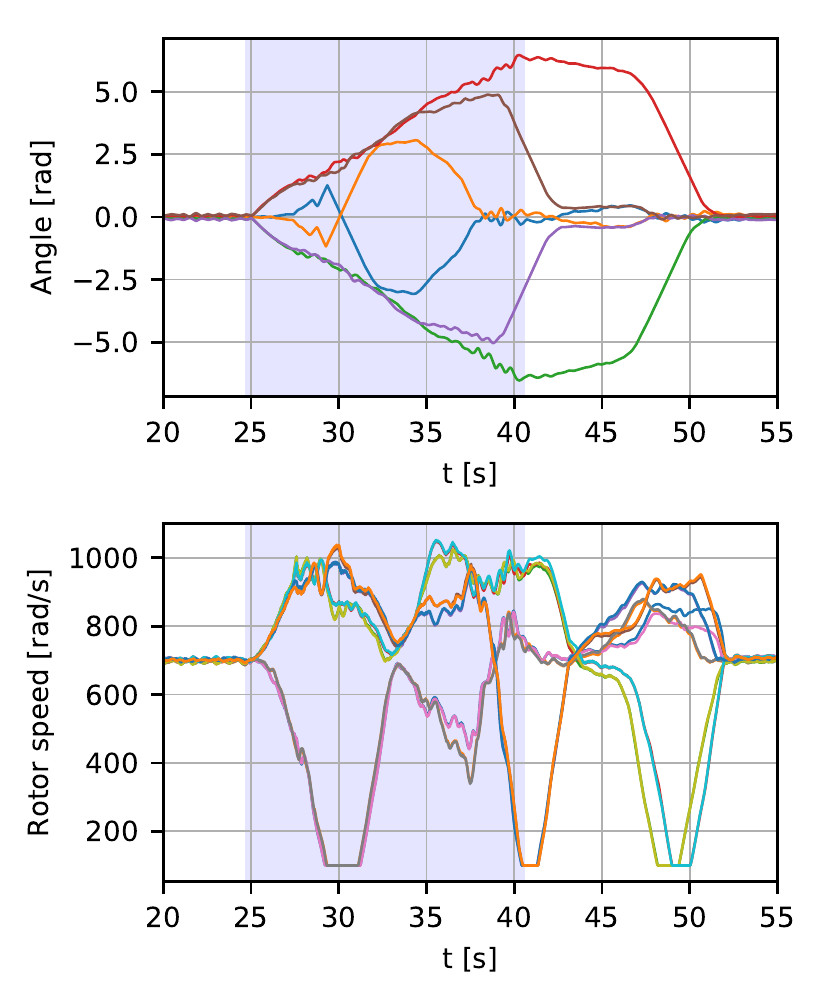}
 \caption{Trajectory (f) actuator commands during a roll flip. The duration of the flip trajectory is highlighted.}
 \label{fig:unwinding_roll_flip}
\end{figure}

\begin{figure}[h]
 \centering
\includegraphics[width=\columnwidth]{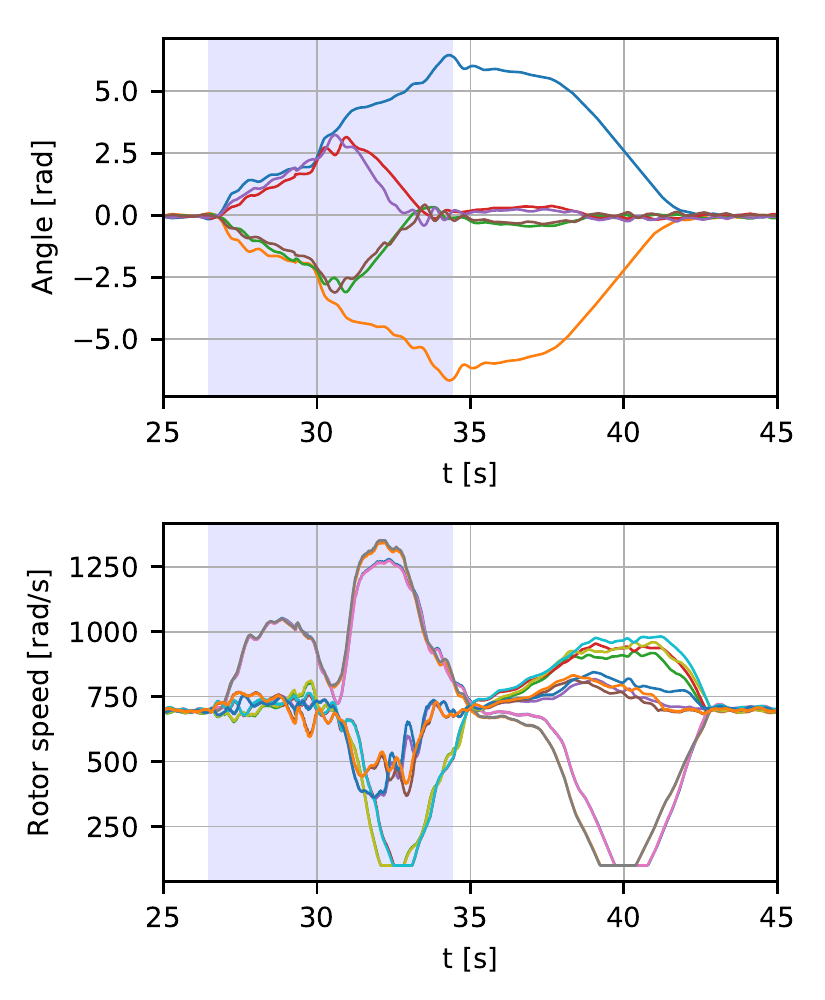}
 \caption{Trajectory (g) actuator commands during a pitch flip. The duration of the flip trajectory is highlighted.}
 \label{fig:unwinding_pitch_flip}
\end{figure}

\subsection{Full pose tracking}
While both control approaches demonstrate stable tracking,  \cref{fig:lqri_pid_pos_att_errors_standard} to \cref{fig:lqri_pid_pos_att_errors_fast} show a better tracking performance under the \ac{PID}
control law. We identify two major sources for the higher tracking error of the \ac{LQRI} controller.
Firstly, all components of the LQRI error state $\vec{e}$ are ultimately dependent on the linear or angular acceleration errors, according to the dynamic matrix $\bm{A}$ from \cref{eq:dynamic_matrix}. Consequently, the controller strongly relies on precise acceleration estimates to perform well. However, due to the introduced delay in the \ac{SG} filter and the numerical differentiation, these estimates may be noisy and inaccurate.
Secondly, the LQRI controller dynamics (especially the feedback linearisation) are based on the system model presented in \cref{sec:model}. That being said, several assumptions made are likely to result in unmodeled forces in the system, e.g. configuration-based airflow interference, and additional dual-rotor effects. We expect that these unmodeled disturbances lead to additional tracking error.
While the integrator action might compensate these effects to a certain extent, it also reduces the ability to track fast trajectories because of the introduced delay.

\subsection{Singularity handling}
The proposed differential allocation is capable of inherently dealing with \textit{kinematic singularities} as described by \cite{morbidi2018energy}. The described optimization ensures finite tilt velocities and thereby allow stable flight, as shown in \cref{fig:singularity_reference_tracking,,fig:lqri_pid_singularity_tracking_error}.

\Cref{fig:singularity_wfi} shows how the proposed \textit{rank reduction singularity} handling procedure reduces the force efficiency index by increasing internal forces. This configuration, while slightly less efficient, allows instantaneous omnidirectional force generation, avoiding the singularity condition.
Results in \cref{fig:singularity_tracking_offsets} do not show an improvement in tracking performance with additional singularity handling. This can be explained by small correction terms of force and torque that are constantly present, so the system is rarely if ever in a true singular state.


\subsection{Unwinding}
We illustrate the advantage of the differential allocation by performing trajectory (a) while commanding four arms to perform a full rotation. This is achieved by starting the experiment with four arms set to $\alpha = 2\pi$, then activating the unwinding task at the start of the trajectory.

\Cref{fig:unwinding_standard_trajectory} illustrates tracking errors during a standard trajectory, comparing with and without unwinding 4 out of 6 arms.
The comparison in \cref{fig:unwinding_standard_trajectory} shows, that despite the full unwinding of 4 arms, the tracking performance is only affected slightly.
\Cref{fig:unwinding_pitch_flip} shows that the unwinding is done sequentially. As the trajectory starts and the pitch increases, two arms are unwound completely in the first half of the trajectory. The opposite two arms are unwound in the second half.
This sequential unwinding of arms results from the optimization of the differential allocation in combination with the given trajectory and has not been specifically designed.

As a comparison, the roll flip control inputs in \cref{fig:unwinding_roll_flip} shows that the optimization triggers the unwinding only after the flip (blue background) has been completed. 
Similarly, during pitch flip (actuator commands shown in \cref{fig:unwinding_pitch_flip}) it can be observed how the 2 arms that are aligned with the y axis perform a full rotation while the remaining 4 arms can already unwind during the second half of the flip. Unwinding for the remaining two arms continues afterwards in steady hover. The delay of unwinding is influenced by the choice of $v_{\dot{\alpha}}, v_{\dot{\omega}}$ and weighting matrix $\bm{W}$, which are selected based on a compromise of responsive unwinding and overall system stability.
The weighting matrix $\bm{W}$ for the differential allocation has been set to be $\text{blkdiag}(\eye{12},k_\alpha\cdot\eye{6})$.

In all three cases, the rotor speeds are ramped down to avoid force inversion. However, because the rotor speeds and therefore the rotor thrusts are not upper bounded in the optimization, the procedure could result in the simultaneous unwinding of more arms than minimally necessary for hover. Additionally, the differential allocation does not incorporate knowledge of the future trajectory and might thus start unwinding, even if it turns out to be unfavourable later on.

\subsection{Complexity and efficiency}

The differential actuator allocation demonstrates prioritization of 6 \ac{DOF} tracking, while optimally exploiting the null space to fulfill secondary tasks. Results show successful kinematic singularity rejection and stable arm unwinding for different trajectories. That being said, the mentioned optimality comes at the cost of higher computational load, since a high dimensional matrix inversion needs to be performed at each iteration. We can justify this computational load with a high performance onboard computer, which completes an iteration of the control loop in \SI{3}{\milli\second} on average.

The additional complexity of the actuation can be justified by comparing to other morphological designs.
The added complexity of tilting rotor groups allows force- and pose-omnidirectionality, enabling the system to track 6\ac{DOF} trajectories with highly dynamic capabilities. This is not possible with regular underactuated multicopters.
Fixed rotor systems that enable full actuation directly select a tradeoff of flight efficiency and omnidirectional force capabilities via their design morphology.

The proposed tiltrotor system takes advantage of both highly efficient flight and omnidirectional capabilities, at the expense of additional weight and actuation complexity. Morphology optimization results presented in \cref{sec:design} quantify these metrics. Verified system performance in experimental flights and the opportunity of null space task prioritization further strengthen the case for the tiltrotor \ac{OMAV}.

\section{Conclusion}
\label{sec:conclusion}

In this paper, the complete system design and optimal control for a novel efficient and versatile tiltrotor \ac{OMAV} has been presented. 
We have open sourced a morphology optimization tool for design of a tiltrotor \acp{OMAV}, and used it to demonstrate the advantages of the system. 
A new controller has been derived and implemented, based on a jerk-level LQRI and a differential actuator allocation. 
We have further integrated secondary tasks to trajectory tracking into the actuation null space, such as maximizing hover efficiency, avoiding cable windup, and singularity handling.
Through various experiments we have compared tracking of the \ac{LQRI} and \ac{PID} controllers, and verified the capabilities of task prioritization in the actuator allocation.

Regarding future work, we intend to improve the system model, and perform a diligent system identification, the importance of which is discussed in \cref{sec:experiments}. We expect that higher model accuracy will improve the \ac{LQRI} performance. The integration of constraints in actuator allocation, and consideration of further secondary tasks, are also the subject of future work. Acceleration estimation has a significant influence on the controller, and will be evaluated in more detail. The controller presented in this paper sets the stage for further optimization over the trajectory horizon, e.g. using a nonlinear \ac{MPC}.

\begin{acks}
This work was supported by funding from ETH Research Grants, the National Center of Competence in Research (NCCR) on Digital Fabrication, NCCR Robotics, and Armasuisse Science and Technology.
\end{acks}

\theendnotes

\bibliographystyle{SageH}
\balance
\bibliography{references.bib}

\newpage
\onecolumn
\appendix

\section{Appendix}
\label{sec:appendix}
\subsection{Morphology Optimization Inertia Calculations}
\label{app:inertia}
We consider a constant core mass $m_{c,const}$ that allows for complete on-board autonomy, an additional core mass $m_a$ to provide tiltrotor actuation for each of $n$ rotor groups, and a mass for each rotor group $m_{\rotor{i}}$. The mass of each rigid arm tube $m_t$ is a function of its length $L$ and length normalized mass $m_{t,norm}$.

\begin{subequations}
    \begin{align}
    m_{c} &= m_{c,const} + n \cdot m_a \\
    m_{t} &= m_{t,norm} \cdot L \\
    m &= m_{c} + \sum_{i=1}^{n}(m_{\rotor{i}} + m_{t}) 
    \end{align}
\end{subequations}

The core inertia is modeled as a solid cylinder centered at $O_\body$ with radius $r_c$ and height $h_c$. 
Rigid tilt arms that connect to propeller groups are modeled as cylindrical tubes, with radii $r_1, r_2$ and length $L$. 
For fairness of comparison, we consider that all systems have single rotor groups with origin $O_{\rotor{i}}$, and a tilt axis $x_{\rotor{i}}$ aligned with the corresponding arm axis. 
The inertia of each rotor group is modeled as a cylinder of radius $r_{\rotor{i}}$ and height $h_{\rotor{i}}$, with inertial values in $y$ and $z$ axes averaged to approximate a system independent of tilt. 
Values are chosen based on components presented in \cref{sec:platform}. 
\begin{subequations}
\begin{align}
    \inertia_{c} &= \text{diag} \Bigg(
    \begin{bmatrix}
    \frac{1}{12}m_{c}(3  {r_c}^2 + {h_c}^2) \\
    \frac{1}{12}m_{c}(3  {r_c}^2 + {h_c}^2) \\ 
    \frac{1}{2}m_{c} {r_c}^2
    \end{bmatrix} \Bigg)
    \\
    \inertia_{t} &= \text{diag} \Bigg(
    \begin{bmatrix}
    \frac{1}{2}m_{t}  {r_c}^2) \\
    \frac{1}{12}m_{t}(3 ({r_1}^2 + {r_2}^2)+ L^2) \\
    \frac{1}{12}m_{t}(3 ({r_1}^2 + {r_2}^2)+ L^2)
    \end{bmatrix} \Bigg)
    \\
    \inertia_{\rotor{i}} &= \text{diag} \Bigg(
    \begin{bmatrix}
    \frac{1}{12}m_{\rotor{i}}(3  {r_{\rotor{i}}}^2 + {h_{\rotor{i}}}^2) \\
    \frac{1}{2}(\frac{1}{12}m_{\rotor{i}}(3  {r_{\rotor{i}}}^2 + {h_{\rotor{i}}}^2) + \frac{1}{2}m_{\rotor{i}}  {r_{\rotor{i}}}^2) \\ 
    \frac{1}{2}(\frac{1}{12}m_{\rotor{i}}(3 {r_{\rotor{i}}}^2 + {h_{\rotor{i}}}^2) + \frac{1}{2}m_{\rotor{i}} {r_{\rotor{i}}}^2)
    \end{bmatrix} \Bigg)
\end{align}
\end{subequations}

Total system inertia is then computed using the parallel axis theorem to express all inertias in $\frame{\body}$. 

\begin{subequations}
    \begin{align}
        \pos_{t} &= \begin{bmatrix} \frac{L}{2} & 0 & 0 \end{bmatrix}^{\top} 
        \\
        \pos_{\rotor{i}} &= \begin{bmatrix} L & 0 & 0 \end{bmatrix}^{\top} 
        \\
        {}_{O_{\body}}\inertia_{t} &=  \inertia_{t} + m_t \norm{\pos_{t}}^2 \eye{3} - \pos_{t} {\pos_{t}}^{\top} 
        \\
        {}_{O_{\body}}\inertia_{\rotor{i}} &=  \inertia_{\rotor{}} + m_{\rotor{i}} \norm{\pos_{\rotor{i}}}^2 \eye{3} - \pos_{\rotor{i}} {\pos_{\rotor{i}}}^{\top} 
        \\
        \rot{\body}{\rotor{i}} &= \rot{z}{}(\gamma_i)\rot{z}{}(\theta_i)\rot{x}{}(\beta_i)
    \end{align}
\end{subequations}
\begin{equation}
    \inertia = \inertia_c + \sum_{i=1}^{n}(\rot{\body}{\rotor{i}}({}_{O_{\body}}\inertia_{t}+{}_{O_{\body}}\inertia_{\rotor{i}}) {\rot{\body}{\rotor{i}}}^{\top})
\end{equation}

\subsection{Angular acceleration dynamics}\label{sec:appendix_jerk}
The dynamics of the angular acceleration error can be expanded to

\begin{align}
    \f{\body}\dot{\vec{e}}_\psi &= 
    \begin{multlined}[t]
    \f{\body}(\dot{\vec{\psi}}_{\world\body}) + [\f{\body}(\dot{\vec{\omega}}_{\world\body})]_\times \bm{R}_{\body\world}\f{\world}\vec{\omega}_{\world\body_\des} 
    + [\f{\body}\vec{\omega}_{\world\body}]_\times\dot{\bm{R}}_{\body\world}\f{\world}\vec{\omega}_{\world\body_\des} \\
    + [\f{\body}\vec\omega_{\world\body}]_\times \bm{R}_{\body\world}\f{\world}(\dot{\vec{\omega}_{\world\body_\des}}) 
    - \dot{\bm{R}}_{\body\world}\f{\world}\vec{\psi}_{\world\body_\des} - \bm{R}_{\body\world}\f{\world}(\dot{\vec{\psi}}_{\world\body_\des})
\end{multlined}\nonumber\\
&=
    \begin{multlined}[t]
        \inertia^{-1}\left(
        \f{\body}(\dot{\vec{\tau}})  
        -\left[[\f{\body}\vec{\omega}_{\world\body}]_\times\vec{r}_{com}\right]_\times\f{\body}\vec{f}
        -[\vec{r}_{com}]_\times\f{\body}(\dot{\vec{f}})\right)\\
        \inertia^{-1}\left(
        -2[\f{\body}\vec{\omega}_{\world\body}]_\times\inertia\f{\body}\vec{\psi}_{\world\body}
        -[\f{\body}\vec{\psi}_{\world\body}]_\times\inertia\f{\body}\vec{\omega}_{\world\body}
        -[\f{\body}\vec{\omega}_{\world\body}]_\times^2\inertia\f{\body}\vec{\omega}_{\world\body}
        \right) \\
        + [\f{\body}\vec{\omega}_{\world\body}]_\times\f{\body}\vec{\psi}_{\world\body} 
        + [\f{\body}\vec{\psi}_{\world\body}]_\times \bm{R}_{\body\world}\f{\world}\vec{\omega}_{\world\body_\des}
        - [\f{\body}\vec{\omega}_{\world\body}]_\times^2 \bm{R}_{\body\world}\f{\world}\vec{\omega}_{\world\body_\des} \\
        + 2[\f{\body}\vec{\omega}_{\world\body}]_\times \bm{R}_{\body\world}\f{\world}\vec{\psi}_{\world\body_\des} - \bm{R}_{\body\world}\f{\world}\vec{\zeta}_{\world\body_\des}
    \end{multlined}\label{eq:error_dynamics_angacc}
\end{align}

\subsection{Linearization}\label{sec:appendix_linearization}
Feedback linearization of angular acceleration error dynamics:

\begin{align}
    \f{\body}(\dot{\vec{\tau}}) &= 
    \begin{multlined}[t]
        \left[[\f{\body}\vec{\omega}_{\world\body}]_\times\f{\body}\vec{r}_{com}\right]_\times\f{\body}\vec{f}
        +[\f{\body}\vec{r}_{com}]_\times\f{\body}(\dot{\vec{f}})
        +2[\f{\body}\vec{\omega}_{\world\body}]_\times\inertia\f{\body}\vec{\psi}_{\world\body}\\
        +[\f{\body}\vec{\psi}_{\world\body}]_\times\inertia\f{\body}\vec{\omega}_{\world\body}
        +[\f{\body}\vec{\omega}_{\world\body}]_\times^2\inertia\f{\body}\vec{\omega}_{\world\body}\\
        + \inertia\left(
        -[\f{\body}\vec{\omega}_{\world\body}]_\times\f{\body}\vec{\psi}_{\world\body} 
        -[\f{\body}\vec{\psi}_{\world\body}]_\times \bm{R}_{\body\world}\f{I}\vec{\omega}_{\world\body_d}
        +[\f{\body}\vec{\omega}_{\world\body}]_\times^2 \bm{R}_{\body\world}\f{I}\vec{\omega}_{\world\body_d} \right)\\
        + \inertia\left(
        - 2[\f{\body}\vec{\omega}_{\world\body}]_\times \bm{R}_{\body\world}\f{I}\vec{\psi}_{\world\body_d} + \bm{R}_{\body\world}\f{I}\vec{\zeta}_{\world\body_d}\right)
        + \inertia
    \begin{bmatrix}
        \bar{u}_4\\\bar{u}_5\\\bar{u}_6
    \end{bmatrix}
\end{multlined}\label{eq:ang_fbl2}
\end{align}

Using \cref{eq:ang_fbl2} in combination with \cref{eq:error_dynamics_angacc} yields a feedback linearized dynamic system of the angular acceleration error dynamics, see \cref{eq:feedback_linearized_system}.

Finally, we can write a linear representation of the error dynamics as follows:
\begin{equation}
    \frac{d}{dt}
    \begin{bmatrix}
        \f{\world}\vec{e}_p\\
        \f{\world}\vec{e}_{p,i}\\
        \f{\world}\vec{e}_{v}\\
        \f{\world}\vec{e}_{a}\\
        \f{\body}\vec{e}_R\\
        \f{\body}\vec{e}_{R,i}\\
        \f{\body}\vec{e}_{\omega}\\
        \f{\body}\vec{e}_{\psi}
    \end{bmatrix}
    =
    \underbrace{
    \begin{bmatrix}
        0 &0 &\eye{}  &0  &0  &0 & 0 & 0 \\
        \eye{} &0 &0  &0   &0  &0 & 0 & 0\\
        0 &0  &0   &\eye{} &0  &0 & 0 & 0\\
        0  &0  &0  &0  &0  &0 & 0 & 0\\
        0 &0 &0 &0 &0  &0 &\eye{} &  0 \\
        0 &0 &0 &0  &\eye{} &0 & 0 & 0 \\
        0 &0 &0 &0 &0  &0 & 0 & \eye{} \\
        0 &0 &0 &0 &0  &0 & 0 & 0 \\
   \end{bmatrix}}_{\bm{A}}
    \begin{bmatrix}
        \f{\world}\vec{e}_p\\
        \f{\world}\vec{e}_{pi}\\
        \f{\world}\vec{e}_{v}\\
        \f{\world}\vec{e}_{a}\\
        \f{\body}\vec{e}_R\\
        \f{\body}\vec{e}_{Ri}\\
        \f{\body}\vec{e}_{\omega}\\
        \f{\body}\vec{e}_{\psi}
    \end{bmatrix}
    +
    \underbrace{
    \begin{bmatrix}
        0 & 0\\
        0 & 0\\
        0 & 0\\
        \eye{} & 0\\
        0 & 0\\
        0 & 0\\
        0 & 0\\
        0 & \eye{}\\
     \end{bmatrix}}_{\bm{B}}
    \begin{bmatrix}
        \bar{u}_1 \\ 
        \bar{u}_2 \\ 
        \bar{u}_3 \\ 
        \bar{u}_4 \\ 
        \bar{u}_5 \\ 
        \bar{u}_6 \\ 
    \end{bmatrix}
    \label{eq:error_dynamics_AB2}
\end{equation}
where $\eye{}$ and $0$ are identity and zero matrices in $\mathbb{R}^{3\times3}$.

\subsection{Stability proof of the proposed LQRI controller}\label{sec:stability_proof}
We prove the stability of the proposed control law \cref{eq:lqr_control} using Lyapunov stability theory. The closed loop system can be written as
\begin{equation}
    \dot{\vec{e}} = \left(\bm{A}-\bm{BK}_{LQRI}\right)\vec{e}
\end{equation}
We define the Lyapunov function:
\begin{equation}
    V(\vec{e})=\vec{e}^T\bm{P}\vec{e}\qquad V(\vec{e})\in\mathbb{R}
\end{equation}
According to Lyapunov stability theory, the system is asymptotically stable, if $\dot{V}(\vec{e})<0,\ \forall\vec{e}\neq0$. The time derivative is computed to be:
\begin{align}
    \dot{V}(\vec{e}) &= \dot{\vec{e}}^T\bm{P}\vec{e} + \vec{e}^T\bm{P}\dot{\vec{e}}\nonumber\\
    &= \vec{e}^T\left( \bm{A}-\bm{BK}_{LQRI} \right)^T\bm{P}\vec{e} + \vec{e}^T\bm{P}\left( \bm{A}-\bm{BK}_{LQRI} \right)\vec{e}\nonumber\\
    &= -\vec{e}^T\left( 
        \bm{Q} +\bm{PBR}^{-1}\bm{B}^T\bm{P}
    \right)\vec{e}
\end{align}
Note that due to $\bm{Q}\succeq0$, $\bm{R}\succ0$, $\bm{P}\succ0$ and the structure of $\bm{B}$, the matrix $\bm{Q} +\bm{PBR}^{-1}\bm{B}^T\bm{P}$ is always positive definite. Therefore, we have $\dot{V}(\vec{e})<0,\ \forall\vec{e}\neq0$ and the closed loop system is asymptotically stable.

Since real system dynamics are nonlinear, the closed loop system needs to be written as follows:
\begin{equation}
    \dot{\vec{e}} = f(\vec{e},\vec{u}=-\bm{K}_{LQRI}\vec{e})
\end{equation}
with the Taylor series of the function $f$ evaluated at the origin:
\begin{equation}
    f(\vec{e},-\bm{K}_{LQRI}\vec{e}) = \bm{A}\vec{e}- \bm{BK}_{LQRI}\vec{e} + \vec{o}|_0(\vec{e}^T\vec{e})
\end{equation}
As above, we can derive the time derivative of the Lyapunov function:
\begin{align}
    \dot{V}(\vec{e}) &= \dot{\vec{e}}^T\bm{P}\vec{e} + \vec{e}^T\bm{P}\dot{\vec{e}}\nonumber\\
    &= -\vec{e}^T\left( \bm{Q}+\bm{PBR}^{-1}\bm{B}^T\bm{P} \right)\vec{e} 
    + 2\vec{e}^T\bm{P}\vec{o}|_0(\vec{e}^T\vec{e})
\end{align}
For the above expression to be $<0$, the following must hold:
\begin{align}
    2\vec{e}^T\bm{P}\vec{o}|_0(\vec{e}^T\vec{e}) &\overset{!}{<}
    \vec{e}^T\left( \bm{Q}+\bm{P}\bm{B}\bm{R}^{-1}\bm{B}^T\bm{P} \right)\vec{e} \nonumber\\
    \norm{2\vec{e}^T\bm{P}\vec{o}|_0(\vec{e}^T\vec{e})} &\overset{!}{<}
    \norm{\vec{e}^T\left( \bm{Q}+\bm{P}\bm{B}\bm{R}^{-1}\bm{B}^T\bm{P} \right)\vec{e}} \nonumber\\
    2\norm{\vec{e}}\norm{\bm{P}}\norm{\vec{o}|_0(\vec{e}^T\vec{e})}
    &\overset{!}{<}
    \norm{\vec{e}}^2\lambda_{min}(\bm{Q}+\bm{P}\bm{B}\bm{R}^{-1}\bm{B}^T\bm{P})\nonumber\\
    \frac{\norm{\vec{o}|_0(\vec{e}^T\vec{e})}}{\norm{\vec{e}}}
    &\overset{!}{<}
    \frac{\lambda_{min}(\bm{Q}+\bm{P}\bm{B}\bm{R}^{-1}\bm{B}^T\bm{P})}{2\norm{\bm{P}}}
\end{align}
where $\lambda_{min}(\cdot)$ denotes the minimum eigenvalue of $(\cdot)$.
Looking at the dynamics in \cref{eq:nonlin}, it is clear that all higher order terms are coming from $\dot{\vec{e}}_{R,\body}$, i.e. 
$
\norm{\vec{o}|_0(\vec{e}^T\vec{e})} = 
\norm{\vec{o}_R|_0(\vec{e}^T\vec{e})}
$. \cite{lee2012robust} shows that 
\begin{equation}
    \norm{\dot{\vec{e}}_{R,\body}} \leq \frac{3}{\sqrt{2}}\norm{\f{\body}\vec{e}_{\omega}}
\end{equation}
This can be extended to 
\begin{align}
    \norm{\dot{\vec{e}}_{R,\body}} &\leq \frac{3}{\sqrt{2}}\norm{\f{\body}\vec{e}_{\omega}}\nonumber\\
    \norm{\f{\body}\vec{e}_{\omega} + \vec{o}_R|_0(\vec{e}^T\vec{e})} &\leq  \frac{3}{\sqrt{2}}\norm{\f{\body}\vec{e}_{\omega}}\nonumber\\
    \norm{\vec{o}_R|_0(\vec{e}^T\vec{e})} -\norm{\f{\body}\vec{e}_{\omega}}  &\leq  \frac{3}{\sqrt{2}}\norm{\f{\body}\vec{e}_{\omega}}\nonumber\\
    \norm{\vec{o}_R|_0(\vec{e}^T\vec{e})} &\leq  \frac{3+\sqrt{2}}{\sqrt{2}}\norm{\f{\body}\vec{e}_{\omega}}
\end{align}
Plugging this into the equation above, we finally get the asymptotic stability condition:
\begin{equation} 
    \label{eq:proof}
    \frac{3+\sqrt{2}}{\sqrt{2}}\frac{\norm{\f{\body}\vec{e}_{\omega}}}{\norm{\vec{e}}}
    \overset{!}{<}
    \frac{\lambda_{min}(\bm{Q}+\bm{PBR}^{-1}\bm{B}^T\bm{P})}{2\norm{\bm{P}}}\\
\end{equation}

Looking at \cref{eq:proof}, we see that a small angular velocity error $\f{\body}\vec{e}_{\omega}$ will make the condition more likely to be fulfilled and thus increase the overall system stability. This can be achieved to some extent by increasing the corresponding weights in the $\bm{Q}$ matrix and thereby changing the objective function for \cref{eq:lqri} accordingly.\\
Real-world experiments have confirmed this behaviour, as generally speaking, higher angular velocity gains lead to less oscillations.\\

Note however, that \cref{eq:proof} is a very conservative and only sufficient condition, meaning that stability might be achieved, even when it is violated.

\subsection{General formulation of the static allocation matrix}\label{sec:appendix_allocation}
For a general tiltrotor MAV with $n$ arms we can write the relation between the body force and torque and the rotor force components as follows:
\begin{equation}
 \begin{bmatrix}\f{\body}\vec{f} \\ \f{\body}\vec{\tau}\end{bmatrix}
 = \bm{A} \tilde{\bm{\Omega}}
 \label{eq:allocation}
\end{equation}
where the rotor force components $\bm{u}$ are defined as the lateral and vertical components of the squared rotor speeds:
\begin{equation}
 \begin{matrix}
  \tilde{\bm{\Omega}} =
  \begin{bmatrix}
  \sin(\alpha_i) \Omega_{i}  \\
  \cos(\alpha_i) \Omega_{i}  \\
  \vdots \\
  \sin(\alpha_n) \Omega_{n} \\
  \cos(\alpha_n) \Omega_{n}
  \end{bmatrix}
  &
  \forall i \in \{1...n\}
 \end{matrix}
\end{equation}
The static allocation matrix $\bm{A}$ can then be written as follows:
\begin{equation}
  \bm{A} = c_f
  \begin{bmatrix}
  \sin(\gamma_i)            & -\sin(\beta_i)\cos(\gamma_i)       & \hdots \\
  -\cos(\gamma_i)           & -\sin(\beta_i)\sin(\gamma_i)       & \hdots \\
  0                         & \cos(\beta_i)                   & \hdots \\
  -s_j c_d \sin(\gamma_i)+l_x\sin(\beta_i)\cos(\gamma_i) & l_x \sin(\gamma_i)+s_ic_d\sin(\beta)\cos(\gamma)  & \hdots \\
  s_i c_d \cos(\gamma_i)+l_x\sin(\beta_i)\sin(\gamma_i)  & -l_x \cos(\gamma_i)+s_ic_d\sin(\beta)\sin(\gamma) & \hdots \\
  -l_x\cos(\beta_i)        & -s_i c_d\cos(\beta_i)            & \hdots
  \end{bmatrix}
\end{equation}
where $\gamma_i$ are the angles between the body $x$ axis and arm $i$ in the body x-y-plane, $\beta_i$ are the angles between the body x-y plane and arm $i$, as illustrated in \cref{fig:arm_angle_definitions}. $c_f$ is the rotor thrust coefficient, $c_d$ is the rotor drag coefficient (relative to $c_f$), and $s_i$ are rotor spin directions of rotors attached to arms $i$.


\end{document}